%% file: main.tex
\documentclass{article}
\usepackage{arxiv}

\usepackage[utf8]{inputenc}
\usepackage[T1]{fontenc}
\usepackage{microtype}
\usepackage{amsmath,amssymb,amsfonts}
\usepackage{booktabs,tabularx,array,multirow}
\usepackage{graphicx,xcolor}
\usepackage{pifont,listings}
\usepackage{flafter}
\usepackage{algorithm,algpseudocode}
\usepackage[numbers,sort&compress]{natbib}
\usepackage{enumitem,caption}
\usepackage{xurl}
\usepackage{tcolorbox}
\usepackage{needspace}
\tcbuselibrary{skins,breakable}
\newtcolorbox{trajbox}[1]{enhanced, breakable, colback=white, colframe=black!45, boxrule=0.5pt, arc=1.5pt,
  left=6pt, right=6pt, top=4pt, bottom=4pt, title={#1}, title after break={#1 (continued)}, fonttitle=\small\bfseries, coltitle=black, colbacktitle=black!7,
  segmentation style={black!35, dashed}, before upper={\interlinepenalty=10000\relax}}
\newenvironment{trajlines}[1]{\par\noindent{\small\textbf{#1}}\par\nobreak\vspace{2pt}\scriptsize\ttfamily\noindent\begin{tabular}{@{}r@{\hspace{6pt}}p{0.86\linewidth}@{}}}{\end{tabular}\par}
\newcommand{\hit}[1]{\textcolor{red!75!black}{#1}}
\usepackage[colorlinks=true,linkcolor=blue!45!black,citecolor=blue!45!black,urlcolor=blue!45!black]{hyperref}

\newcommand{\papertitle}{Multi-Task Self-Evolving Harnesses with the Agent as Its Own Optimizer}
\hypersetup{pdftitle={\papertitle},pdfauthor={Qiankai Xu}}
\renewcommand{\shorttitle}{\papertitle}
\renewcommand{\headeright}{}
\renewcommand{\undertitle}{}
\makeatletter
\renewcommand{\@maketitle}{%
  \vbox{%
    \hsize\textwidth
    \linewidth\hsize
    \vskip 0.1in
    \@toptitlebar
    \centering
    {\LARGE\sc \@title\par}
    \@bottomtitlebar
    \vskip 0.06in
    \begin{tabular}[t]{c}\bf\@author\end{tabular}\par
    \vskip 0.22in
  }%
}
\makeatother

\setlist{itemsep=0.15em,topsep=0.3em,leftmargin=1.6em}
\newcolumntype{Y}{>{\raggedright\arraybackslash}X}
\newcommand{\headstack}[1]{\begin{tabular}[b]{@{}c@{}}#1\end{tabular}}
\newcommand{\yes}{\checkmark}
\newcommand{\no}{\ding{55}}
\newcommand{\agent}{\mathcal{A}}
\newcommand{\train}{\mathrm{tr}}
\newcommand{\test}{\mathrm{ho}}

\title{Multi-Task Self-Evolving Harnesses\\ with the Agent as Its Own Optimizer}
\author{Qiankai Xu\textsuperscript{1}}
\date{}

\begin{document}
\maketitle
{\renewcommand{\thefootnote}{1}\footnotetext{Independent Researcher. Correspondence to: Qiankai Xu \textless qiankaixu6@gmail.com\textgreater.}}

\begin{abstract}
A harness is the code around a language model agent that organizes prompts, calls tools, manages context, and controls execution. As models grow stronger, recent work has begun to let agents improve their own harnesses, a line of work known as self-evolving harnesses. In most existing methods, a separate proposer running on a human-designed harness modifies the solver's harness, and a separate harness is evolved for each benchmark. Real-world tasks come from many domains, so both the evolution and the evaluation of a harness should cover a diverse range of tasks. We propose a framework close to recursive self-improvement: the same frozen model, on the same version of the harness, first solves tasks as the solver and then, as the proposer, reads the complete run records and directly edits the harness that runs it. Each evolution batch draws tasks from five benchmarks in different domains. To measure generalization, training and held-out tasks are strictly separated, and we additionally evaluate on five out-of-distribution benchmarks not used in the first stage. The design borrows from deep learning training and has two stages, multi-task pretraining and continual training. Starting from a 49-line seed harness, the harnesses evolved by GLM-5.3 and GPT-5.6 Sol gain about 5 points on the in-distribution benchmarks. They are on par with Claude Code and Codex on these benchmarks, and the GPT-5.6 Sol harness is also on par with Codex on five out-of-distribution benchmarks. The weaker model of each family benefits less: GLM-5.3-Flash gains less than 2 points, and the harness evolved by GPT-5.6 Luna scores below the seed. Continued evolution on Claw-Eval in the second stage further improves GPT-5.6 Sol and GLM-5.3-Flash on its multimodal tasks. We also analyze where the gains come from and the mechanisms that emerged during evolution, including output truncation, history compaction, completion checks, independent review, and visual inspection. Code and the evolved harnesses are available at \href{https://github.com/xqkkk/self-evolving-harness}{this link}.
\end{abstract}

\input{figures/pipeline}
\input{sections/introduction}
\input{sections/related_work}
\input{sections/method}
\input{sections/experiments}
\input{sections/analysis}
\input{sections/discussion}

\bibliographystyle{unsrtnat}
{\small\bibliography{references}}

\input{sections/appendix}
\end{document}

%% file: figures/pipeline.tex
\begin{figure}[t]
\centering
\includegraphics[width=\linewidth]{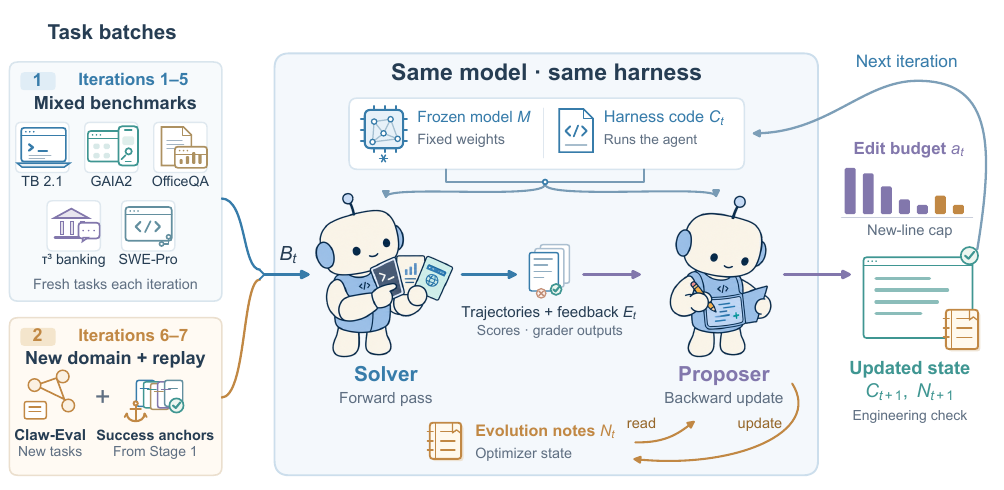}
\caption{Overview of the framework. The solver and the proposer are the same agent, run in separate sessions with the same frozen model $M$ and the same harness, which consists of the code $C_t$ and the evolution notes $N_t$. The solver runs the batch $B_t$ (new tasks from five benchmarks in Iterations~1--5; new Claw-Eval tasks plus success anchors, tasks already solved in the first stage, in Iterations~6--7) and produces the run records $E_t$: trajectories, scores, and grader outputs. The proposer reads these records and the notes, modifies the harness code within the budget of new lines $a_t$, and updates the notes. The new version is used in the next iteration once it passes the engineering checks.}
\label{fig:pipeline}
\end{figure}

%% file: sections/introduction.tex
\section{Introduction}
\label{sec:introduction}

The performance of a language model agent depends both on the model itself and on the control code around it. This code decides what information the model sees at each step, which tools it can call, what to keep when the conversation grows long, and when a task counts as finished. It is usually called the harness. Work such as ReAct, CodeAct, and SWE-agent shows that the design of action formats and tool interfaces substantially changes what an agent can do~\citep{react,codeact,sweagent}, and with the same model, changing only the harness can also produce large performance differences~\citep{metaharness,ahe}. Most harnesses are still written by engineers, and developers of frontier models build dedicated harnesses such as Codex and Claude Code for their own models.

Self-evolving harnesses aim to hand this engineering work to the agent. Recent work mostly follows the same basic framework: a solver works on the tasks of a benchmark with the current harness and produces trajectories, a proposer reads these trajectories and modifies the solver's harness, and the loop repeats. The proposer is usually driven by a strong model and runs on a fixed, human-designed harness. The outer loop often adds further human-designed steps, such as a dedicated agent that organizes trajectories~\citep{ahe,harnessx}.

\subsection{Self-Improvement Framework with Minimal Human Priors}
\label{sec:intro-self}

If the goal is to improve the harnesses used by frontier agents, this framework has four limitations:
\begin{itemize}
  \item \textbf{Models of the proposer and solver.} Harness improvements often help weaker models more~\citep{metaharness,harnessx,harnessupdating}, so it is unsurprising that a strong model can build an effective harness for a weaker one. The more interesting question is whether a frontier model can improve its own harness beyond human-designed ones.
  \item \textbf{Harnesses of the proposer and solver.} The proposer runs on a fixed, human-designed harness. No matter how much the solver's harness improves, the proposer keeps working the same way and never benefits from these improvements.
  \item \textbf{Scope of edits.} Methods such as AHE, HarnessX, and Self-Harness decide beforehand which components, edit types, or configuration points the proposer is allowed to change~\citep{ahe,harnessx,selfharness}, which leaves no room for mechanisms outside these predefined categories.
  \item \textbf{Human-designed outer pipelines.} Predefined component structures and agents that organize trajectories mainly compensate for the weaknesses of current models, and they may become unnecessary as models improve~\citep{oeo,bitterlesson}.
\end{itemize}
To address these four limitations, we adopt a setting close to recursive self-improvement (RSI)~\citep{good1965,godelmachine,stop}: a frontier model starts from a minimal initial harness and modifies the harness that runs it. The same agent plays two roles. As the solver it solves tasks, and as the proposer it reads the complete run records of the previous batch and edits the harness directly (Figure~\ref{fig:pipeline}). Both roles use the same frozen model and the same version of the harness. The proposer runs on the harness it is modifying, so every edit to the harness also changes how the proposer works. The setting follows two principles. \textbf{First, minimize human-designed priors}, so that the mechanisms in the harness emerge from the model's own experience. \textbf{Second, give the model maximal freedom to modify the harness}, with no restriction on what can be changed or how.

This setting reflects our long-term vision. Once models are capable enough, harnesses will no longer need human design. A model can start from scratch and keep improving its harness based on the tasks it receives, its own attempts, and the feedback it gets, so that the harness fits both the model and its tasks while remaining general and capable.

\subsection{Generalization of Harness Updates}
\label{sec:intro-general}

A harness is ultimately used on tasks it has not seen, so the improvements found by evolution must generalize beyond the training tasks. Prior studies find that evolution tends to memorize training tasks or overfit to a particular benchmark, and that the gains shrink markedly, or even vanish, on new tasks or out-of-distribution benchmarks~\citep{rrsi,modularrsi,harnesscompass,harnessevolve,selfevolvingcoding}. To test generalization and to evolve one harness that serves many domains, we take three measures:
\begin{itemize}
  \item \textbf{Train/held-out separation.} The tasks of each benchmark are split into disjoint training and held-out sets, and the effect of evolution is measured only on held-out tasks.
  \item \textbf{Out-of-distribution evaluation.} We also evaluate on five benchmarks never used in the first stage.
  \item \textbf{Multi-benchmark mixing.} Most prior work evolves a separate harness for each benchmark (Appendix~\ref{app:related}). In practice, an agent faces tasks from many domains, so one harness has to work across many benchmarks. Each of our training batches therefore draws tasks from five benchmarks in different domains, and an edit has to work for all task types at once~\citep{caruana1997}.
\end{itemize}

\subsection{Design Borrowed from Model Training}
\label{sec:intro-training}

Our experimental design borrows from the way model parameters are trained. SkillOpt trains skill documents in this way~\citep{skillopt}, and we apply the same view to the entire harness source code: the source code corresponds to the parameters, and each rewrite of the source by the proposer corresponds to a parameter update. The design therefore follows three common practices in training: mixing data from multiple sources~\citep{pile}, decaying the learning rate~\citep{sgdr}, and replaying old data during continual training~\citep{mccloskey1989,rolnick2019,ibrahim2024}. Appendix~\ref{app:loop} gives the full correspondence (Table~\ref{tab:analogy} and Figure~\ref{fig:training}).

\begin{figure}[t]
\centering
\includegraphics[width=\linewidth]{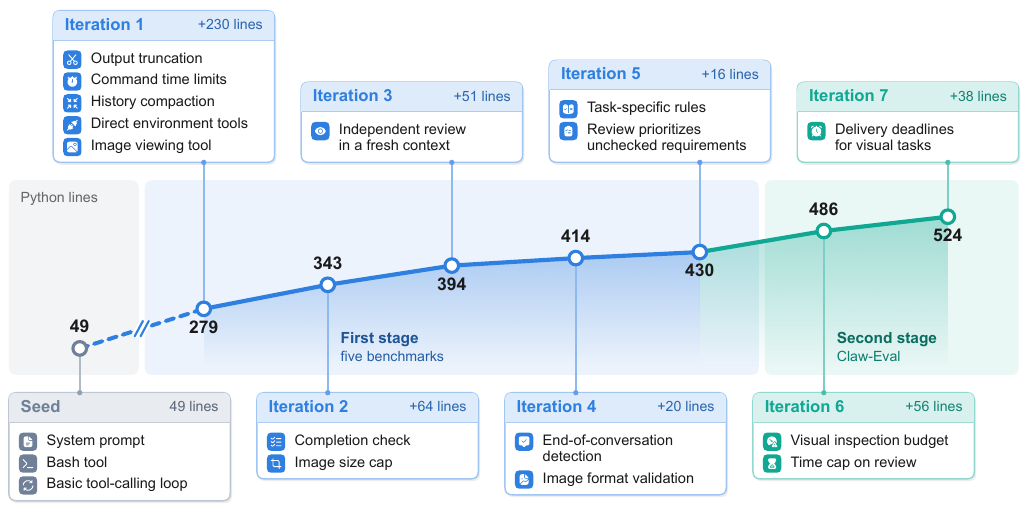}
\caption{Evolution of the GPT-5.6 Sol harness in Run~1. The curve gives the number of Python lines of the harness after each iteration, with the vertical scale broken between Seed and Iteration~1. Each card lists the main changes made in one iteration and the net change in the number of lines, and the Seed card lists what the seed contains.}
\label{fig:evolution-timeline}
\end{figure}

\bigskip

In our experiments, we run the first stage on four models and the second stage on GPT-5.6 Sol and GLM-5.3-Flash. In the first run of GPT-5.6 Sol, the harness grows from 49 lines to 430 in the first stage and 524 in the second, gaining output truncation, history compaction, and independent review, and later a visual inspection budget (Figure~\ref{fig:evolution-timeline}). On the held-out tasks of the five evolution benchmarks, the harnesses evolved by GLM-5.3 and GPT-5.6 Sol score about 5 points above the seed harness on average and are on par with their production harnesses (Table~\ref{tab:main-results}). GLM-5.3-Flash gains only 1.93 points, and GPT-5.6 Luna ends below the seed (Section~\ref{sec:capability}). On five out-of-distribution benchmarks never used in the first stage, the harness evolved by GPT-5.6 Sol is on par with Codex and averages about 12.5 points above the seed harness (Table~\ref{tab:ood-results}). Most of this gain comes from BrowseComp-Plus, and the other four benchmarks stay close to the seed. In the second stage, further evolution on Claw-Eval raises the scores of GPT-5.6 Sol and GLM-5.3-Flash on its multimodal tasks by 10.42 and 16.98 points (Figure~\ref{fig:claw-heldout}), mainly through changes to image viewing.

Our main contributions are as follows:
\begin{enumerate}
  \item \textbf{Self-evolving framework.} A self-evolving harness framework with minimal human priors: the same model solves tasks on the same harness and modifies the harness that runs it. Both evolution and evaluation mix benchmarks from multiple domains, and the design borrows from model training, with two stages of multi-task pretraining and continual training.
  \item \textbf{Experiments.} We compare the seed harness, the production harness commonly used with each model, and the evolved harness on four models, measure generalization on held-out tasks and out-of-distribution benchmarks, and evaluate GPT-5.6 Sol and GLM-5.3-Flash on a new domain.
  \item \textbf{Analysis.} Using the code changes and trajectories, we analyze which kinds of runs the gains come from, the mechanisms shared by all models and those specific to each, how the gains relate to model capability, and the drawbacks of the design.
\end{enumerate}

%% file: sections/related_work.tex
\section{Related Work}
\label{sec:related}

\paragraph{Self-improving agents and harness evolution.}
For language models, G{\"o}del Agent lets an agent modify its own logic at run time, DGM and SICA let coding agents rewrite their own code bases, and Live-SWE-agent starts with only a bash tool and writes new tools while it solves a task~\citep{godelagent,dgm,sica,liveswe}. Since 2026 the focus has shifted to harnesses~\citep{metaharness,ahe,harnessx}, and later work has introduced designs such as candidate banks, population merging, module-wise evolution, and validation gates~\citep{harnessbank,darwinx,modularrsi,evolvenet}. Appendix~\ref{app:related} compares representative methods item by item (Table~\ref{tab:related}) and discusses further related work.

%% file: sections/method.tex
\section{Method}
\label{sec:method}

\subsection{Problem Setup}
\label{sec:problem}

Let $M$ denote the frozen base model. At the end of iteration $t$, the harness state is $H_t=(C_t,N_t)$, where $C_t$ is the entire source code that runs the agent, including the system prompt, tool definitions and implementations, context handling, and execution control, and $N_t$ denotes the evolution notes stored alongside the code (Section~\ref{sec:training-view}). $H_0$ is the seed harness.

The harness runs on a fixed execution interface $\mathcal{K}$, which calls the model, executes commands, and records the entire run. Given a task $x$, one run of the agent is
\begin{equation}
    \tau=\agent_{\mathcal{K}}(M,H_t,x),
    \label{eq:agent}
\end{equation}
where $\tau$ contains the model replies, tool calls, command outputs, and final artifacts. The grader of benchmark $k$ returns $(r,v)=V_k(x,\tau)$, where $r$ is the task score and $v$ is the test log or grading explanation.

\subsection{Solver, Proposer, and Seed Harness}
\label{sec:roles}
\label{sec:principles}
\label{sec:seed}

Following the two principles of Section~\ref{sec:intro-self}, the framework makes three design choices: the solver and the proposer are the same agent, the outer program adds no analysis step, and the seed harness is minimal.

The \textbf{solver} receives benchmark tasks and solves them, and the \textbf{proposer} receives an update task $I_t$ that asks it to improve the harness. Both roles are played by the same agent with the same $M$ and the same $H_t$, and each runs in a fresh session.

The outer program only dispatches tasks and collects run records. There is no step that organizes trajectories or summarizes errors. The update task provides three kinds of material: the source code and evolution notes of the current harness; the complete run records of the previous batch, including trajectories, task scores, and grader outputs, for both successful and failed tasks and without any summarization or selection; and, starting from the second iteration of each stage, the run record of the previous update session. The update prompt is short (Appendix~\ref{app:instruction} gives the full text). It tells the proposer that the modified harness will be evaluated on unseen tasks and on entirely different benchmarks, so it should look for improvements that generalize. It also says that the proposer is free to add or delete files and introduce any component, and that the notes are passed on with the harness to the next round. Beyond this, the prompt does not say which records to read first, how to identify problems, where to start, or which part to change. During evolution, the proposer cannot access the internet apart from a few sites such as package registries, so it can rely only on the model's own knowledge and these run records to improve the harness.

The seed $H_0$ is a single 49-line Python file (Appendix~\ref{app:seed}) containing a system prompt of a few sentences, a bash tool, and a minimal loop: call the model, execute the tool calls in its reply and append the results to the history, and stop once the model replies without tool calls. The seed has no output truncation, command timeouts, context compaction, or error recovery. Whether these are needed, and how to implement them, is left to evolution. The framework also does not divide the harness into components or restrict the scope of edits: apart from the execution interface $\mathcal{K}$, any part of any file in the harness can be modified, and files can be created or deleted.

\subsection{One Iteration: Rollout, Update, and Check}
\label{sec:iteration}

Iteration $t+1$ starts from $H_t$ and consists of three steps: rollout, update, and check (Figure~\ref{fig:pipeline}; Algorithm~\ref{alg:loop} in Appendix~\ref{app:loop}). In the rollout, the solver runs the task batch $B_t$ of the iteration with $H_t$, once per task, producing the records $E_t=\{(x,\tau_x,r_x,v_x):x\in B_t\}$. In the update, the outer program combines $E_t$ with the other materials described in Section~\ref{sec:roles} into the update task $I_t$ and passes it to the same agent:
\begin{equation}
    \widetilde H_{t+1}
    =\operatorname{Extract}\!\left(\agent_{\mathcal{K}}(M,H_t,I_t)\right),
    \label{eq:update}
\end{equation}
where $\operatorname{Extract}$ retrieves the modified code and notes at the end of the session. In the check, the new version must pass a few engineering checks $G_t$: the edits change only the harness itself, there is at least one change besides the notes, the number of new lines does not exceed $a_t$, and the harness completes a full task run with a mock model:
\begin{equation}
    H_{t+1}=
    \begin{cases}
      \widetilde H_{t+1}, & G_t(\widetilde H_{t+1},H_t)=1,\\
      H_t, & \text{otherwise}.
    \end{cases}
    \label{eq:accept}
\end{equation}
If the update session fails, for example because its edit does not pass the check or the session runs out of time, the outer program keeps $H_t$, starts a repair session, and then runs the update again (Appendix~\ref{app:instruction}).

\subsection{Two Stages: Multi-Task Pretraining and Continual Training}
\label{sec:stages}

\textbf{Stage 1: multi-task pretraining (Iterations~1--5).} There are $K=5$ benchmarks, each split in advance into a training set $D_k^{\train}$ and a disjoint held-out set $D_k^{\test}$. Each iteration takes $b=12$ previously unused tasks from each training set to form a batch of 60 tasks:
\begin{equation}
    B_t=\bigcup_{k=1}^{K}B_{t,k},\qquad
    B_{t,k}\subset D_k^{\train},\qquad |B_{t,k}|=b,\qquad
    B_{t,k}\cap B_{u,k}=\varnothing\ (t\neq u).
    \label{eq:batch}
\end{equation}
Together, the five iterations use each of the 60 training tasks of every benchmark exactly once, and the held-out sets are never run during evolution. Appendix~\ref{app:batch-size} compares evolution runs that differ in batch size.

\textbf{Stage 2: continual training on a new domain (Iterations~6--7).} Starting from Iteration~5, the training tasks switch to tasks $B_t^{\mathrm{new}}$ from a new domain, mixed with a small number of old tasks $A_t$:
\begin{equation}
    B_t=B_t^{\mathrm{new}}\cup A_t,\qquad t\in\{5,6\}.
    \label{eq:new-domain-batch}
\end{equation}
The two iterations use different new tasks. $A_t$ is drawn from the training tasks solved in the first stage, spread evenly across the five benchmarks. We call these tasks success anchors. They are rerun with the current harness, and their results are given to the proposer together with the records of the new tasks. This follows the practice in continual learning of mixing in old data, which is meant to prevent catastrophic forgetting.

\subsection{Evolution Notes and Edit Budget}
\label{sec:training-view}

\paragraph{Evolution notes.}
In each iteration, the proposer first reads the existing notes and then, based on the new batch of run records, writes down which edits worked as expected, which earlier judgments need correction, and what to try next. The design is inspired by the first moment in Adam, which accumulates past gradients with exponential weighting so that each update reflects both the current gradient and the direction of earlier updates~\citep{adam,adamw}.

\paragraph{Edit budget.}
The budget $a_t$ of each iteration limits the number of lines added between two versions of the source code, i.e., $A(C_t,\widetilde C_{t+1})\le a_t$. Deleted lines and edits to the notes do not count toward the budget. The first-stage budget follows a cosine learning rate schedule~\citep{sgdr} and decreases from 500 to 100 lines, giving budgets of 500, 441, 300, 159, and 100 lines for the five iterations (Appendix~\ref{app:loop} gives the formula). When the second stage enters the new domain, the budget is raised back to 200 lines and then lowered to 100 lines, following the practice in continual pretraining of first raising the learning rate again and then decaying it~\citep{ibrahim2024}. We introduced the budget after early exploratory runs without one. In those runs, the harness accumulated patterns that hold only for individual tasks, and the proposer still made large edits in the last iterations (Appendix~\ref{app:cross-model}). In the experiments of this paper, the two GLM models came close to the budget several times, adding 498 and 496 of the 500 lines in Iteration~1, whereas the two GPT models stayed below it.

%% file: sections/experiments.tex
\section{Experiments}
\label{sec:experiments}

\subsection{Experimental Setup}
\label{sec:setup}

\paragraph{Benchmarks.} We select benchmarks that are difficult and far from saturated for current models, and that are as diverse as possible: they either cover entirely different kinds of tasks or each span many kinds of tasks. The first stage uses five benchmarks that differ in interaction style, task length, and grading: Terminal-Bench~2.1~\citep{terminalbench}, GAIA2~\citep{gaia2}, OfficeQA Pro~\citep{officeqa}, $\tau^3$-Bench banking~\citep{tauknowledge}, and SWE-Bench~Pro~\citep{swebenchpro}. Each has 60 training tasks and 29 to 100 held-out tasks. To test whether the evolved harness generalizes beyond the benchmarks it was evolved on, we also evaluate on five out-of-distribution (OOD) benchmarks never used in the first stage: SWE-bench Verified~\citep{swebench,swebenchverified}, DeepSWE~\citep{deepswe}, BrowseComp-Plus~\citep{browsecompplus}, APEX-Agents~\citep{apexagents}, and Claw-Eval~\citep{claweval}. Appendix~\ref{app:benchmarks} lists the task type, the number of tasks, and the number of evaluation repeats of each benchmark. Time limits and grading follow the official settings of each benchmark.

\input{sections/main_table}

\input{sections/ood_table}

\begin{figure}[t]
\centering
\includegraphics[width=\linewidth]{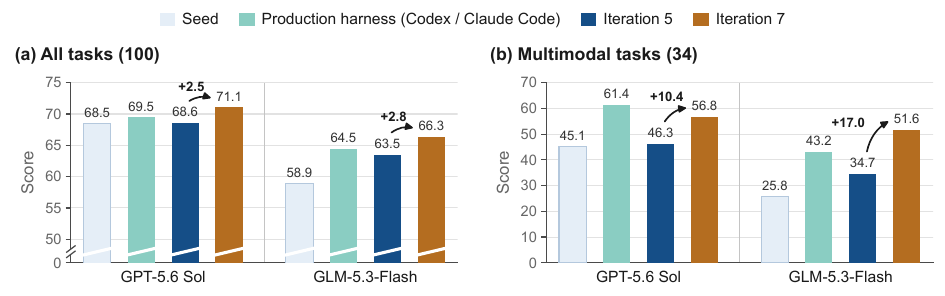}
\caption{Results on the Claw-Eval held-out tasks for GPT-5.6 Sol and GLM-5.3-Flash. The production harness is Codex for GPT-5.6 Sol and Claude Code for GLM-5.3-Flash. Scores are the mean official score multiplied by 100, over all 100 tasks in (a) and over the 34 multimodal tasks in (b). Arrows mark the change from Iteration~5 to Iteration~7. The vertical axis of (a) is broken between 0 and 50.}
\label{fig:claw-heldout}
\end{figure}

\paragraph{Models and baselines.} The first stage is run separately on four models: GLM-5.3, GPT-5.6 Sol, GLM-5.3-Flash, and GPT-5.6 Luna. The GLM models use max reasoning effort and the GPT models use high. All four start from the same seed and share the same training tasks, budgets of new lines, and update prompt. For each model we compare three harnesses: the seed harness (Seed); the production harness commonly used with the model, built by engineers, which is Codex\footnote{\url{https://github.com/openai/codex}} for the GPT models and Claude Code\footnote{\url{https://github.com/anthropics/claude-code}} for the GLM models; and Iteration~5, the harness evolved by the end of the first stage. The production harness runs the same model at the same reasoning effort. To keep the comparison fair, runs that fail because of infrastructure problems, such as sandbox creation failures, network interruptions, or provider errors, are rerun. Appendix~\ref{app:cost} reports the tokens, time, and API cost of all evolution and evaluation runs.

\paragraph{Repeated runs and confidence intervals.}
\label{sec:repeats}
Because a single evolution run is subject to randomness, the first stage of GPT-5.6 Sol is run three times independently, denoted Run~1 to Run~3. All three start from the same seed and use the same model, reasoning effort, and outer loop, with identical training tasks at every iteration. We report the mean and standard deviation over the three runs, and the out-of-distribution evaluation, the second stage, and the analysis of mechanisms in Section~\ref{sec:shared} use the harness from Run~1. For the score differences compared in this paper we give 95\% confidence intervals, obtained by a paired bootstrap over tasks (Appendix~\ref{app:ci}).

\paragraph{Preventing reward hacking.} To prevent reward hacking, such as bypassing the task to obtain answers directly or exploiting loopholes in grading~\citep{metrrewardhacking,hal}, we apply the same two safeguards to all three harnesses in every evaluation. Beforehand, every task description ends with a fixed requirement that forbids obtaining answers from published solutions, upstream patches, or evaluation files, and reference answers and grading tests are placed in the environment only after the agent has finished. Afterward, a script scans all evaluation trajectories and flags suspicious behavior, and other agents then check the flagged records one by one. A record confirmed as cheating scores 0. No cheating was confirmed in any evaluation record reported in this paper.

\subsection{Main Results}
\label{sec:main-results}

\paragraph{In-distribution results.}
Table~\ref{tab:main-results} reports the scores of the four models on the held-out tasks of the five in-distribution benchmarks. Starting from the 49-line seed, the Iteration~5 harnesses evolved by GLM-5.3 and GPT-5.6 Sol score 5.26 and 4.95 points above Seed on average, and both confidence intervals exclude zero. The Iteration~5 harness of GLM-5.3-Flash scores 1.93 points above Seed, with an interval that includes zero. All three models gain most on Terminal-Bench~2.1. The three runs of GPT-5.6 Sol give similar results: their Iteration~5 harnesses average 60.71, 60.99, and 60.23, with a standard deviation of 0.38 points, and each is more than 4.5 points above Seed. Terminal-Bench~2.1 and SWE-Bench~Pro improve in all three runs, while the other three benchmarks rise in some runs and fall in others.

For these three models, Iteration~5 is also on par with the production harness built by engineers: their average scores differ by only 0.1 to 1.1 points (Appendix~\ref{app:ci} gives the confidence intervals). GPT-5.6 Luna is the exception, with Iteration~5 below Seed (Section~\ref{sec:capability}). Appendix~\ref{app:regressions} discusses the individual benchmarks on which Iteration~5 or the production harness falls below Seed.

\paragraph{Out-of-distribution results.}
\label{sec:ood-results}
Table~\ref{tab:ood-results} reports the results of GPT-5.6 Sol on the five out-of-distribution benchmarks. Iteration~5 averages 12.47 points above Seed (95\% confidence interval 8.63 to 16.32) and is almost equal to Codex (70.93 vs.\ 71.23). Most of this gain comes from BrowseComp-Plus. On this benchmark, Seed often exceeds the input limit because command outputs are too long, whereas Iteration~5 truncates overly long outputs (Section~\ref{sec:truncation}), and the score rises from 34 to 90. On the other four benchmarks Iteration~5 is close to Seed: it averages 1.58 points higher, with a confidence interval that includes zero. DeepSWE is 5.66 points higher, and each of the other three changes by less than 1 point. On these four benchmarks, which were never used in evolution, we therefore observe no drop relative to Seed. How much the evolved mechanisms help depends on how often Seed runs into the problems that they solve (Section~\ref{sec:gains}).

\subsection{Continual Training on Claw-Eval}
\label{sec:claw-results}

\paragraph{Setup.}
The second stage is run with GPT-5.6 Sol and GLM-5.3-Flash under the same setup: starting from its own Iteration~5, each model runs two consecutive iterations on Claw-Eval, yielding Iteration~7. Each iteration uses 40 Claw-Eval training tasks, different for the two iterations, mixed with 20 success anchors, i.e., training tasks the model already solved in the first stage (Section~\ref{sec:stages}).

\paragraph{Results.}
Figure~\ref{fig:claw-heldout} compares Iteration~5 and Iteration~7 of the two models, with Seed and the production harness as references. Over all 100 tasks, Iteration~7 scores 2.54 and 2.81 points above Iteration~5 for the two models. The gain comes from the 34 multimodal tasks, on which the two models improve by 10.42 and 16.98 points. Few training tasks in the first stage require viewing images, and the two models changed how the harness views images in the second stage (Section~\ref{sec:visual-evolution}).

\paragraph{Regression on the first-stage benchmarks.}
The new tasks of the second stage all come from Claw-Eval, so we also checked how Iteration~7 of GPT-5.6 Sol performs on the five first-stage benchmarks. Its average score is 2.69 points below Iteration~5, and GAIA2 falls from 47 to 38 points. The GAIA2 drop is caused by a rule added in the last update. The rule makes one category of GAIA2 tasks impossible to complete, and the success anchors read by that update happened to include no task of this category. So even with a few old tasks replayed, a new edit can still harm a kind of task that the replay does not cover. Without any replay of old tasks, such conflicts would very likely be more frequent, an effect similar to catastrophic forgetting. Appendix~\ref{app:regressions} gives the details.

%% file: sections/main_table.tex
\begin{table}[t]
\centering\small
\caption{Held-out scores (success rates, \%) on the five in-distribution benchmarks. The first stage of GPT-5.6 Sol is run three times, and its Iteration~5 row gives the mean and standard deviation over the three runs. The average weights the five benchmarks equally. $\Delta$ is the change of the average relative to Seed, followed by its 95\% confidence interval from a paired bootstrap over tasks. The highest score of each model is in bold.}
\label{tab:main-results}
\renewcommand{\arraystretch}{1.04}
\setlength{\tabcolsep}{2.5pt}
\begin{tabular*}{\linewidth}{@{\extracolsep{\fill}}llrrrrrrl@{}}
\toprule
\textbf{Model} & \textbf{Harness} & \textbf{TB 2.1} & \textbf{GAIA2} & \textbf{\headstack{OfficeQA\\Pro}} & \textbf{\headstack{$\tau^3$\\banking}} & \textbf{\headstack{SWE-Bench\\Pro}} & \textbf{Average} & \textbf{$\Delta$ [95\% CI]} \\
\midrule
\multirow{3}{*}{GLM-5.3} & Seed & 50.57 & \textbf{52.00} & 58.22 & 45.95 & \textbf{62.00} & 53.75 & -- \\
 & Claude Code & 61.15 & 51.00 & 63.24 & \textbf{59.01} & 60.00 & 58.88 & +5.13 {\scriptsize[+1.0, +9.3]} \\
 & Iteration~5 & \textbf{63.22} & 49.00 & \textbf{65.07} & 56.76 & 61.00 & \textbf{59.01} & +5.26 {\scriptsize[+1.4, +9.1]} \\
\midrule
\multirow{4}{*}{GPT-5.6 Sol} & Seed & 66.67 & 43.00 & \textbf{66.44} & 51.35 & 51.00 & 55.69 & -- \\
 & Codex & 78.16 & \textbf{53.00} & 63.70 & 45.95 & 57.00 & 59.56 & +3.87 {\scriptsize[$-$0.4, +8.2]} \\
 & Iteration~5 & \textbf{84.67} & 42.67 & 65.30 & \textbf{52.25} & \textbf{58.33} & \textbf{60.64} & +4.95 {\scriptsize[+1.5, +8.6]} \\[-3pt]
 &  & {\scriptsize$\pm$\,3.51} & {\scriptsize$\pm$\,5.13} & {\scriptsize$\pm$\,2.85} & {\scriptsize$\pm$\,3.12} & {\scriptsize$\pm$\,3.79} & {\scriptsize$\pm$\,0.38} & \\
\midrule
\multirow{3}{*}{GLM-5.3-Flash} & Seed & 45.98 & \textbf{51.00} & 59.59 & \textbf{51.35} & 58.00 & 53.18 & -- \\
 & Claude Code & \textbf{60.92} & 48.00 & 57.99 & 43.24 & \textbf{61.50} & 54.33 & +1.15 {\scriptsize[$-$2.6, +4.9]} \\
 & Iteration~5 & 56.32 & \textbf{51.00} & \textbf{60.96} & 47.30 & 60.00 & \textbf{55.12} & +1.93 {\scriptsize[$-$2.0, +6.0]} \\
\midrule
\multirow{3}{*}{GPT-5.6 Luna} & Seed & 55.17 & 43.00 & 54.79 & 28.38 & 51.00 & 46.47 & -- \\
 & Codex & \textbf{73.56} & \textbf{51.00} & \textbf{56.85} & \textbf{32.43} & \textbf{60.00} & \textbf{54.77} & +8.30 {\scriptsize[+4.1, +12.6]} \\
 & Iteration~5 & 55.17 & 28.00 & \textbf{56.85} & 12.16 & 34.00 & 37.24 & $-$9.23 {\scriptsize[$-$13.8, $-$4.8]} \\
\bottomrule
\end{tabular*}
\end{table}

%% file: sections/ood_table.tex
\begin{table}[t]
\centering\small
\caption{Held-out scores of the three harnesses with GPT-5.6 Sol on the five out-of-distribution benchmarks. Scores on APEX-Agents and Claw-Eval are the mean official score multiplied by 100, and success rates (\%) on the other benchmarks. The average and $\Delta$ are defined as in Table~\ref{tab:main-results}. The highest score in each column is in bold.}
\label{tab:ood-results}
\renewcommand{\arraystretch}{1.04}
\setlength{\tabcolsep}{3pt}
\begin{tabular*}{\linewidth}{@{\extracolsep{\fill}}lrrrrrrl@{}}
\toprule
\textbf{Harness} & \textbf{\headstack{SWE-bench\\Verified}} & \textbf{DeepSWE} & \textbf{\headstack{BrowseComp-\\Plus}} & \textbf{\headstack{APEX-\\Agents}} & \textbf{Claw-Eval} & \textbf{Average} & \textbf{$\Delta$ [95\% CI]} \\
\midrule
Seed & 78.00 & 62.26 & 34.00 & 49.56 & 68.49 & 58.46 & -- \\
Codex & \textbf{80.00} & 62.26 & \textbf{93.00} & \textbf{51.40} & \textbf{69.50} & \textbf{71.23} & +12.77 {\scriptsize[+8.6, +16.9]} \\
Iteration~5 & 78.00 & \textbf{67.92} & 90.00 & 50.16 & 68.56 & 70.93 & +12.47 {\scriptsize[+8.6, +16.3]} \\
\bottomrule
\end{tabular*}
\end{table}

%% file: sections/analysis.tex
\section{Analysis}
\label{sec:analysis}

This section first explains where the gains in score come from, and then describes the evolved mechanisms, the differences between models, and the drawbacks of the design. Appendices~\ref{app:cross-model}, \ref{app:offtheshelf}, and~\ref{app:traces} give the per-model changes, a comparison with production harnesses, and example trajectories.

\subsection{Where the Gains Come From}
\label{sec:gains}

\begin{figure}[t]
\centering
\includegraphics[width=\linewidth]{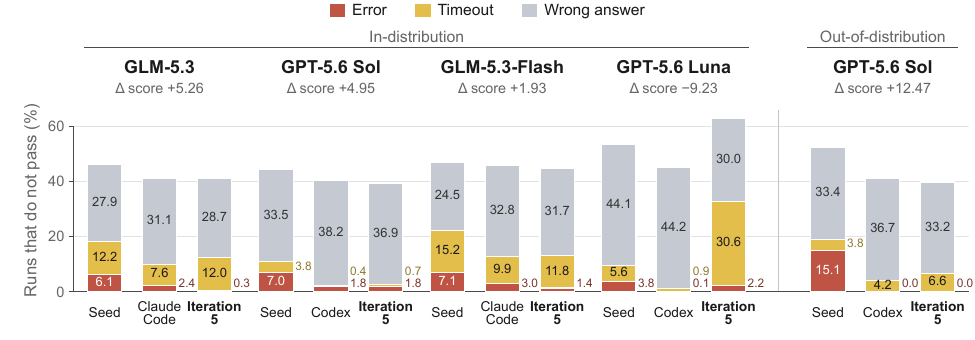}
\caption{Outcomes of the held-out runs that do not pass, under Seed, the production harness (Codex or Claude Code), and Iteration~5. Each column stacks the shares of runs that end with an error, a timeout, or a wrong answer, and the remaining runs pass. The share of a thin segment is printed beside the column. Shares are averaged over the five benchmarks of each group. $\Delta$ score is the score change of Iteration~5 relative to Seed. On the in-distribution benchmarks, Iteration~5 of GPT-5.6 Sol is the mean of its three runs.}
\label{fig:outcomes}
\end{figure}

A run that does not pass ends in one of three ways. \textbf{Error}: the run ends abnormally, for example because a request exceeds the model's input limit or a model call fails. \textbf{Timeout}: the run is still going when the task time limit is reached. \textbf{Wrong answer}: the run ends normally, but the result is wrong. Figure~\ref{fig:outcomes} compares the shares of these three kinds of runs under Seed, the production harness, and Iteration~5.

\paragraph{Errors nearly disappear.}
With the seed harness, 4\% to 7\% of the runs of every model end with an error. After evolution, this share falls to about 2\% or less, on par with the production harnesses. On the out-of-distribution benchmarks, 15\% of the Seed runs end with an error and none of the Iteration~5 runs do.

\paragraph{Wrong answers do not decrease.}
Once the error that stopped a run is fixed, the run continues until it gives a result, and the result can be right or wrong. So only some of the runs that no longer end with an error pass, and the others end with a wrong answer or a timeout. This is why, for the three models whose scores improve, the share of wrong answers stays about the same as with Seed or rises. For GPT-5.6 Sol, almost all of the gain comes from runs that ended with an error or a timeout under Seed, both in and out of distribution. GLM-5.3 is different: more than half of its gain comes from runs that ended with a wrong answer under Seed and pass after evolution. This shows that, besides fixing basic problems such as errors, harness evolution can also improve the ability of the model to solve tasks (Appendix~\ref{app:outcomes}).

\paragraph{Timeouts depend on the model.}
For GPT-5.6 Sol, timeouts fall from 4\% to below 1\% on the in-distribution benchmarks. They decrease somewhat for GLM-5.3-Flash and stay about the same for GLM-5.3. On the out-of-distribution benchmarks, the timeouts of GPT-5.6 Sol rise from 4\% to 7\%. Almost all of the added timeouts are runs that ended with an error or a wrong answer under Seed, mainly on Claw-Eval (Section~\ref{sec:visual-evolution}). For GPT-5.6 Luna, timeouts rise from 6\% to 31\%, which is the main reason its score ends below Seed (Section~\ref{sec:capability}).

\subsection{Evolved Mechanisms}
\label{sec:shared}

Starting from the same seed, all four models added output truncation, command time limits, and history compaction within the first two iterations, and three of them also added a completion check before finishing (Figure~\ref{fig:evolution-timeline} shows the iterations of GPT-5.6 Sol). These mechanisms all solve problems that the seed ran into on the first batch of training tasks.

\paragraph{Output truncation and time limits.}
\label{sec:truncation}
The seed appends command output to the conversation history unchanged and places no limit on how long a command can run. A single overly long output makes the next request exceed the model's input limit, and the task aborts. All four models added output truncation in Iteration~1, passing only part of an overly long output to the model, and set time limits for commands within the first two iterations, each with its own thresholds. With GPT-5.6 Sol, BrowseComp-Plus was affected most (Figure~\ref{fig:mechanism-firing}): its corpus is large, a broad \texttt{rg} search can print tens of millions of characters, and the seed aborted on 65 tasks for this reason. Receiving truncated output, Iteration~5 narrows its searches and reads candidate files one at a time, and it answers 57 of these 65 tasks correctly.

\paragraph{History compaction.}
The conversation history grows as a task proceeds and eventually exceeds the model's input limit. All four models added history compaction in Iteration~1, which shrinks the history when it approaches the limit, in one of two ways: GPT-5.6 Sol and GPT-5.6 Luna delete earlier messages, while the two GLM models keep every message and only shorten long tool outputs in earlier ones. Compaction lets long tasks continue. In tasks that involve interacting with a user, however, the user's requests reach the model through tool results, and deleting earlier messages can delete these requests as well (Appendix~\ref{app:regressions}). With GPT-5.6 Luna, this problem affects almost all interactive tasks (Section~\ref{sec:capability}).

\paragraph{Completion check.}
The seed ends a task as soon as the model replies without calling any tool, but such a reply is neither written to a file nor sent to the user, so the task often ends with nothing delivered. GLM-5.3, GPT-5.6 Sol, and GLM-5.3-Flash all added a completion check in Iteration~2: when it receives such a reply, the harness sends it back to the model and asks the model to confirm that the required outputs have been delivered before the task ends. GPT-5.6 Luna never added this mechanism.

\label{sec:differences}Beyond these shared mechanisms, the harnesses differ considerably across models, and Iteration~5 ranges from 221 to 1086 lines of code (Appendix~\ref{app:cross-model}).

\paragraph{Independent review.}
Only GPT-5.6 Sol turned the completion check into an independent review, in Iteration~3, whereas the checks of the two GLM models always stay in the original conversation. The notes of Iteration~3 argue that when the same session rechecks its own result with the whole history of its work still in the conversation, the second check tends to repeat the reasoning of the first. For tasks that need no interaction, the harness therefore clears the conversation, keeps only the original task and the file system, and lets the model review the result independently in a fresh context. Interactive tasks keep their history to avoid repeating actions already taken.

\paragraph{Visual inspection.}
\label{sec:visual-evolution}
In the first stage, only GPT-5.6 Sol added an image-viewing tool, which lets the model look at images directly. Multimodal tasks require viewing images or video frames repeatedly, yet an answer or a file must be delivered within 600 seconds, and Iteration~5 of GPT-5.6 Sol failed to deliver in time on 22 of the 34 multimodal held-out tasks because of repeated viewing. In the second stage, it limited the number of images that can be viewed per reply and per task and added a delivery deadline, and Iteration~7 no longer times out on these tasks. Iteration~5 of GLM-5.3-Flash had no way to view images. In the second stage it added an image-viewing tool and corrected how it estimates the input length taken up by images.

\subsection{Evolution Gains and Model Capability}
\label{sec:capability}

\paragraph{Gains differ across models.}
If we measure a model's capability by its score with the production harness, the stronger model of each family gains more from evolution (Table~\ref{tab:main-results}).

\paragraph{Why updating the harness is hard.}
Modifying one's own harness is much harder than solving a single task. The proposer has to read 60 run records from five domains, decide which failures are caused by the harness, write correct code, and anticipate how an edit affects the other domains, and nothing in this process is validated. Both of the weaker models went wrong at this step.

\paragraph{GPT-5.6 Luna.}
After GPT-5.6 Luna modified its harness, its average score fell 9.23 points below Seed, mainly because of history compaction. The compaction it added in Iteration~1 is only guaranteed to keep the system prompt and the task description, so the rest of the history can be deleted, and the next three iterations reduced the history kept after compaction step by step from 480,000 to 48,000 characters. Runs had become longer because compaction deleted information the model needed, but GPT-5.6 Luna interpreted this as the model repeating useless attempts and kept reducing the retained history (Appendix~\ref{app:regressions}). In tasks that involve interacting with a user, the user's requests were deleted as well, and the model inspected the environment again and again until the task timed out (Figure~\ref{fig:outcomes}).

\paragraph{GLM-5.3-Flash.}
Iteration~5 of GLM-5.3-Flash likewise removed most of the errors and some of the timeouts, but runs with wrong answers rose by about 7 percentage points (Figure~\ref{fig:outcomes}), and defects in the implementation canceled part of the gain (Appendix~\ref{app:regressions}). The second stage was different. The problems in the new domain, Claw-Eval, were concentrated and obvious: almost all multimodal tasks in the first training batch timed out, GLM-5.3-Flash added an image-viewing tool in the very next iteration, and its second-stage gain is comparable to that of GPT-5.6 Sol (Section~\ref{sec:claw-results}).

These results suggest that the evolution task defined in this paper demands considerable model capability, especially when the causes of failure are spread across several domains. We expect the gains from this framework to grow as models become more capable.

\subsection{Drawbacks of Simplicity and Freedom}
\label{sec:costs}

Our framework is deliberately simple: the outer loop has no analysis or validation step, and the scope of edits is unrestricted. This design has three drawbacks.

\paragraph{Local fixes with side effects.}
Each round, the proposer reads only one batch of run records and focuses its edits on the most common failures in that batch, so side effects on other tasks may be noticed only when they appear again in later batches. For example, the history compaction that GPT-5.6 Sol added for long tasks deletes the customer's requests on $\tau^3$-Bench banking, which puts the score of its first run on this benchmark below Seed (Appendix~\ref{app:regressions}).

\paragraph{Acceptance without validation.}
A new version is accepted as long as it runs. The framework has no validation set and does not compare the scores of the old and new versions, so whether an edit helps depends almost entirely on the model's own judgment. This is where harness evolution differs most from deep learning training. A sufficiently small step along the negative gradient is guaranteed to lower the loss on the current batch, whereas here the update direction is decided by the model, may be wrong, and is not checked by any validation step. The last edit of GPT-5.6 Sol in the second stage is an example (Section~\ref{sec:claw-results}). The weaker the model, the greater this risk~\citep{oeo,harnessupdating}. The framework still lacks a way to validate each update at an acceptable cost.

\paragraph{Freedom without diversity.}
Although the scope of edits is unrestricted, the edits made by the four models concentrate on a few kinds of mechanisms. Components that the update prompt mentions as examples, such as skills, memory, and subagents, never appear in the evolved harnesses. One remedy is to keep multiple versions, as DGM does, so that evolution proceeds along several paths at once~\citep{dgm}.

%% file: sections/discussion.tex
\section{Discussion and Conclusion}
\label{sec:discussion}

\paragraph{Limitations.}
Our experiments cover four models. Only GPT-5.6 Sol repeats the first stage and is evaluated on the out-of-distribution benchmarks, and only two models run the second stage, which involves a single new domain. Whether the conclusions extend to more models and domains remains to be tested. Some benchmarks have few held-out tasks, and several of the differences we report have confidence intervals that include zero, in particular those between the evolved and the production harnesses. We have not yet ablated each design choice separately, so the experiments test the effectiveness of the pipeline as a whole.

\paragraph{Conclusion.}
We propose a new self-evolving harness framework in which the same frontier model solves tasks on the same harness and directly modifies the harness that runs it, with no human-designed analysis step in the outer loop and no restriction on the scope of edits. Evolution tasks mix benchmarks from multiple domains, and the process borrows from model training, with two stages of multi-task pretraining and continual training. Starting from a 49-line seed, the harnesses evolved by GLM-5.3 and GPT-5.6 Sol are on par with Claude Code and Codex, and for GPT-5.6 Sol this also holds on the out-of-distribution benchmarks. Continued evolution on a new domain in the second stage further improves GPT-5.6 Sol and GLM-5.3-Flash on the multimodal tasks of that domain. Validating each update at an acceptable cost and letting evolution proceed along several paths at once are promising ways to further improve harness self-evolution.

\subsubsection*{Broader Impact Statement}

This work lets an agent modify the harness that runs it, which carries two risks: the agent may remove safety restrictions written into the harness, or bypass the task during evaluation to obtain answers directly. All our safeguards are in parts that the agent cannot modify: the fixed execution interface, the check that accepts only edits to the harness itself, isolated sandboxes, limited network access for the proposer, and a scan of evaluation trajectories for reward hacking (Sections~\ref{sec:method} and~\ref{sec:setup}). When deploying a self-evolving harness, we recommend likewise keeping safeguards outside the harness and reviewing the evolved code before use.

%% file: sections/appendix.tex
\appendix
\raggedbottom

\section{Execution Interface and Seed Harness Code}
\label{app:seed}

The execution interface $\mathcal{K}$ offers the harness only two calls. \texttt{model.chat(messages, tools)} sends a list of messages and tool definitions to the model and returns its reply and tool calls, and \texttt{world.exec(command)} runs a shell command in the task environment. The interface itself provides no tools, does not handle images, and does not connect to the tool servers of a task environment. Features such as exposing environment tools directly to the model or viewing images were implemented by the models themselves on top of these two calls. The update prompt states the input limit of the model (Listing~\ref{lst:instruction}).

The complete code of the seed harness (Section~\ref{sec:roles}) is shown below.

\begin{lstlisting}[caption={Complete code of the seed harness (49 lines).},label={lst:seed}]
SYSTEM_PROMPT = (
    "You are an autonomous agent working inside a Linux container.\n"
    "Use the bash tool to inspect the environment and complete the task.\n"
    "Never cheat: do not obtain the answer by searching the internet or any "
    "similar shortcut; solve the task yourself.\n"
    "When the task is done, reply with a final message and no tool calls."
)

TOOLS = [
    {
        "type": "function",
        "function": {
            "name": "bash",
            "description": "Run a bash command in the task container and return its output.",
            "parameters": {
                "type": "object",
                "properties": {
                    "command": {"type": "string", "description": "The command to run."}
                },
                "required": ["command"],
            },
        },
    }
]

async def run(instruction, model, world):
    messages = [
        {"role": "system", "content": SYSTEM_PROMPT},
        {"role": "user", "content": instruction},
    ]
    while True:
        reply = await model.chat(messages, tools=TOOLS)
        messages.append(reply.message)
        if not reply.tool_calls:
            return
        for call in reply.tool_calls:
            if call.name == "bash":
                r = await world.exec(str(call.arguments.get("command", "")))
                parts = [r.stdout]
                if r.stderr:
                    parts.append("[stderr]\n" + r.stderr)
                parts.append(f"[exit {r.return_code}]")
                output = "\n".join(p for p in parts if p)
            else:
                output = f"unknown tool: {call.name}"
            messages.append(
                {"role": "tool", "tool_call_id": call.id, "content": output}
            )
\end{lstlisting}

\section{Update Task Prompt}
\label{app:instruction}

The update prompt that the proposer receives at each iteration is shown below. It is given as an ordinary task to the agent running on the current harness, with the current harness's own system prompt. In the first iteration, this is the system prompt in Listing~\ref{lst:seed}. The last paragraph is the fixed requirement appended to every task description (Section~\ref{sec:setup}). \texttt{<a\_t>} is the budget of new lines for the iteration (Section~\ref{sec:training-view}). Apart from this budget, the prompt is identical in both stages.

\begin{lstlisting}[language={},numbers=none,caption={Full text of the update prompt.},label={lst:instruction}]
You are improving the agent harness that runs you. The goal is a harness that generalizes: it will be evaluated on tasks it has not seen, some from these same benchmarks and some from entirely different ones. Memorizing specific tasks or their solutions is therefore worthless, and so is hard-coding wordings or rules that restate a single trial, such as a fixed clarification question or a dictated report phrasing. Look for the general, high-level improvements the evidence points to; guidance should teach a method and leave the wording to the model.

`/app/harness_workspace` hold the harness. `/app/harness_kernel` is the fixed kernel that loads the workspace; read it for context, but it is not editable in this task. `/app/eval_trials/` holds the trial records of this harness's past runs. Edits count only under `/app/harness_workspace`. 

You are free to reshape the harness however you see fit -- add, split, or delete files, and introduce whatever components you judge it needs: skills, tools, prompts, memory, subagents, context management, planning, or anything of your own design. Over many rounds this should grow into a complete, well-rounded harness that handles all kinds of tasks, so explore boldly rather than keep reworking one spot.

This is one round of a loop. The harness you leave here is the one that runs the next batch of tasks and the one that runs this editing task again on those trials -- round after round. 

The workspace carries a permanent notebook, `harness_workspace/evolve_notes.md`, that travels with the harness from round to round; bring it up to date before you finish. It can record whether earlier edits delivered what was expected, why or why not, and whether their gains generalized to this round's tasks; your reading of the current evidence; what you changed and what you expect it to improve; which ideas have proven ineffective; and whatever else you judge worth recording. Feel free to rewrite or delete notes that have become outdated or disproven. Do not try to record the specifics of any one task in the notebook -- the same task will not appear again. 

Files written under `/logs/agent` in any task container, this editing one included, are collected into that trial's `agent/` directory, and those trials are what the next editing round sees. That is how the model, now and during the coming tasks, chooses what to keep for future rounds: a trajectory in its own format, notes on what went wrong, whatever it judges worth keeping.

The model API accepts at most 258,400 input tokens per request; over that, the request fails with `context_length_exceeded`.

Acceptance checks run after you finish: changes must be confined to `harness_workspace/` and must not be empty aside from `evolve_notes.md`, and `cd /app && python -m harness_kernel.smoke /app/harness_workspace` must print `SMOKE OK`. This round you are allowed to add at most <a_t> new lines; edits to `evolve_notes.md` do not count, and neither do deleted lines. The smoke command counts and checks this for you. Leave your changes uncommitted.
Never cheat: do not obtain the answer from any source that carries it rather than solving the task -- the upstream fix or its pull request, a published solution or test file, a dataset copy of the answer, or an oracle/evaluation artifact left in the environment. Reading ordinary documentation and library source is fine. Solve the task yourself.
\end{lstlisting}

\paragraph{Repair sessions.}
An update session counts as failed if it runs out of time, exceeds the input limit while reading the run records, crashes, leaves the code unchanged, or produces an edit that does not pass the check. Failures caused by the infrastructure are not repaired, and the session is simply rerun. After an update session fails, the outer program starts a repair session. The repair session is again an ordinary task for the agent: it receives the current harness and the record of the failed session, but no run records of the training tasks, and its prompt asks it to repair the harness so that the same kind of failure cannot happen again, changing only what the repair needs. The repair can add at most 100 new lines and must pass the same check.

The first repair session runs on the current harness. If it also fails, each further repair session is run by the harness of one iteration earlier (the harness being repaired is still the current one), since a broken harness may be unable to repair itself, and at most three repair sessions are tried. After a successful repair, the outer program runs the update session again from the repaired harness with the full update task, so the edit of that iteration consists of the repair and the new update. In the experiments reported in the main text, this happened once. In the fifth update session of GLM-5.3-Flash, a test script the model had written never terminated, and the session ran out of time. The repair session added a time limit on commands that shrinks as the run goes on, and the rerun update produced Iteration~5. The second stage does not use repair sessions: a failed update session is simply run again, up to three times.

\section{Evolution Loop and Training View}
\label{app:loop}

Algorithm~\ref{alg:loop} gives the evolution loop of Section~\ref{sec:iteration} in pseudocode, and Figure~\ref{fig:training} depicts the two stages of evolution as training. Table~\ref{tab:analogy} lists the correspondence between training and harness evolution that guided the design.

\begin{algorithm}[!htb]
\caption{The harness self-evolution loop shared by both stages}
\label{alg:loop}
\begin{algorithmic}[1]
\Require frozen model $M$ and execution interface $\mathcal{K}$, seed $H_0$, task batches $\{B_t\}_{t=0}^{6}$ and budgets $\{a_t\}_{t=0}^{6}$ of the iterations
\Ensure harness states $H_1,\ldots,H_7$ and all run records
\For{$t=0,\ldots,6$}
  \State $E_t\gets\varnothing$
  \For{$x\in B_t$} \Comment{solver, parallelizable}
    \State $\tau_x\gets\agent_{\mathcal{K}}(M,H_t,x)$
    \State $(r_x,v_x)\gets V_{k(x)}(x,\tau_x)$
    \State $E_t\gets E_t\cup\{(x,\tau_x,r_x,v_x)\}$
  \EndFor
  \State build the update task $I_t$ from $E_t$, the record of the previous update session, $H_t$, and $a_t$
  \State $\widetilde H_{t+1}\gets\operatorname{Extract}(\agent_{\mathcal{K}}(M,H_t,I_t))$ \Comment{proposer}
  \State check with Eq.~\eqref{eq:accept} to obtain $H_{t+1}$
\EndFor
\State freeze $H_5$ and $H_7$ and evaluate them on the corresponding held-out tasks
\end{algorithmic}
\end{algorithm}

\input{figures/training_view}

\input{figures/analogy_table}

The budget of new lines of iteration $t$ in the first stage is
\begin{equation}
    a_t=\operatorname{round}\!\left[
    a_{\min}+\frac{a_{\max}-a_{\min}}{2}
    \left(1+\cos\frac{\pi t}{T_{\mathrm{pre}}-1}\right)\right],
    \qquad t=0,\ldots,T_{\mathrm{pre}}-1,
    \label{eq:schedule}
\end{equation}
with $a_{\max}=500$, $a_{\min}=100$, and $T_{\mathrm{pre}}=5$.

\section{Additional Related Work}
\label{app:related}

\paragraph{Harness gains and model capability.}
The gains from harness improvements depend on model capability. On Terminal-Bench~2.0, Meta-Harness improves Claude Haiku~4.5 more than Claude Opus~4.6~\citep{metaharness}, and HarnessX likewise observes that the weakest task model benefits most~\citep{harnessx}. Harness Updating Is Not Harness Benefit finds that the strongest tier of models gains less from updates than mid-tier models~\citep{harnessupdating}. OEO finds that a sufficiently strong optimizer model performs better when it organizes the optimization process itself than when it follows a predefined pipeline~\citep{oeo}. In these works, the harness is improved by humans or by a stronger model. In our framework, the model evolves its own harness.

\paragraph{Evolution tasks and evaluation.}
SICA, Meta-Harness, and HarnessDev let one harness receive feedback from multiple benchmarks or datasets, all of the same task type~\citep{sica,metaharness,harnessdev}. SEAGym and Evo-Bench find that both the diversity of experience sources and the task domain affect the outcome of evolution~\citep{seagym,evobench}. Simple Baselines and Rethinking the Evaluation of Harness Evolution stress that evolution methods should be compared with simple methods under the same budget~\citep{simplebaselines,rethinkingeval}.

\paragraph{Optimization-style evolution.}
TextGrad and GEPA update prompts with textual feedback and trajectory reflection, respectively~\citep{textgrad,gepa}. SkillOpt uses bounded edits and a textual learning rate budget when training skill documents, which directly inspired our training analogy~\citep{skillopt}. The concurrent RRSI also gradually reduces the number of edits a candidate can contain across iterations~\citep{rrsi}. Adaptive Auto-Harness and Continual Harness study continual evolution, targeting streams of heterogeneous incoming tasks and embodied tasks without environment resets, respectively~\citep{adaptiveharness,continualharness}.

\input{sections/related_table}

\section{Benchmarks}
\label{app:benchmarks}

Table~\ref{tab:benchmarks} lists the ten benchmarks of Section~\ref{sec:setup} with their task counts. Among the five first-stage benchmarks, Terminal-Bench~2.1 consists of long-horizon engineering tasks, such as software builds and system configuration, carried out in a terminal~\citep{terminalbench}. GAIA2 involves handling asynchronous requests in a simulated mobile-app environment that pushes new messages over time~\citep{gaia2}. OfficeQA Pro requires retrieving evidence from nearly a century of U.S. Treasury bulletins and performing numerical reasoning~\citep{officeqa}. $\tau^3$-Bench banking involves multi-turn conversations with a simulated user to handle banking requests under business rules~\citep{tauknowledge}. SWE-Bench~Pro asks the agent to fix real issues in large code repositories~\citep{swebenchpro}. The table also gives the task types of the five out-of-distribution benchmarks. The corpus of BrowseComp-Plus, about 100,000 web pages, is placed in the task environment as files.

\begin{table}[!htb]
\centering\small
\caption{Benchmarks and task counts. Benchmarks with few held-out tasks are evaluated repeatedly and averaged: each of the 29 Terminal-Bench~2.1 tasks is run three times, written as 29$\times$3. For OfficeQA Pro and $\tau^3$-Bench banking we take two complete runs, written as 73$\times$2 and 37$\times$2. The out-of-distribution benchmarks have no first-stage training tasks, and the Claw-Eval training tasks are used only in the second stage.}
\label{tab:benchmarks}
\renewcommand{\arraystretch}{1.15}
\begin{tabularx}{\linewidth}{@{}lYrrr@{}}
\toprule
\textbf{Benchmark} & \textbf{Task type} & \textbf{Total} & \textbf{Train} & \textbf{Held-out} \\
\midrule
\multicolumn{5}{@{}l}{\textit{In-distribution benchmarks}} \\
Terminal-Bench 2.1 & Long-horizon engineering tasks in a terminal & 89 & 60 & 29$\times$3 \\
GAIA2 & Asynchronous interaction in dynamic app environments & 800 & 60 & 100 \\
OfficeQA Pro & Treasury document retrieval and numerical reasoning & 133 & 60 & 73$\times$2 \\
$\tau^3$-Bench banking & Multi-turn service dialogue with a simulated user & 97 & 60 & 37$\times$2 \\
SWE-Bench Pro & Issue resolution in code repositories & 731 & 60 & 100 \\
\midrule
\multicolumn{5}{@{}l}{\textit{Out-of-distribution benchmarks}} \\
SWE-bench Verified & Issue resolution in Python repositories & 500 & -- & 100 \\
DeepSWE & Feature development in multi-language repositories & 113 & -- & 53 \\
BrowseComp-Plus & Deep-research question answering over a fixed corpus & 830 & -- & 100 \\
APEX-Agents & Professional work in banking, consulting, and law & 480 & -- & 100 \\
Claw-Eval & Service orchestration, multimodal understanding, and multi-turn dialogue & 300 & 80 & 100 \\
\bottomrule
\end{tabularx}
\end{table}

\section{Confidence Intervals}
\label{app:ci}

The 95\% confidence intervals in this paper are computed with a paired bootstrap over tasks. In each round, we resample the tasks of every benchmark with replacement, as many as the benchmark has, and score the two harnesses on the same resampled tasks, where the score of a task is the mean over its runs. We then recompute the score of each benchmark and the average over the benchmarks, which gives one difference between the two harnesses. We repeat this 10{,}000 times, and the interval runs from the 2.5th to the 97.5th percentile of these differences. The interval describes how much the difference would vary if a different set of held-out tasks of the same kind were used. Table~\ref{tab:ci-comparisons} lists the differences compared in the main text with their confidence intervals.

\input{sections/ci_table}

\section{Resource Use}
\label{app:cost}

Tables~\ref{tab:cost} and~\ref{tab:cost-eval} report the tokens, running time, and API cost of all runs in this paper. Table~\ref{tab:cost} covers the evolution runs, split into solving the training tasks and the update sessions, and Table~\ref{tab:cost-eval} covers the evaluation on held-out tasks. Runs that were replaced after an infrastructure failure are not counted. Tokens and time come from the record of each run. The cost is computed from the token counts at the official API prices of each model (standard tier). Per million tokens of input, cached input, and output, these are \$4.00, \$0.40, and \$20.00 for GPT-5.6 Sol, \$0.20, \$0.02, and \$1.20 for GPT-5.6 Luna, \$1.40, \$0.26, and \$4.40 for GLM-5.3, and \$0.15, \$0.03, and \$0.50 for GLM-5.3-Flash. For the two GPT models, input that is not read from the cache is charged at the cache write price (\$5.00 and \$0.25).

One complete first-stage run costs from \$23 for GLM-5.3-Flash to \$650 for the first run of GPT-5.6 Sol, and the second stage costs \$5 and \$161 for these two models. Solving the training tasks accounts for nearly all of the cost of evolution, and the update sessions take less than 10\% of the input tokens and of the cost. One first-stage run uses fewer tokens than evaluating one harness once on the held-out tasks of the five benchmarks. The tasks of a batch run in parallel, and from start to end the first stage usually takes 9 to 14 hours and the second stage about 5 hours.

The three runs of GPT-5.6 Sol use similar numbers of input tokens, yet their costs differ by a factor of three, because the evolved harnesses differ in how well they use the provider's prompt cache. The seed harness only appends to the end of the history, so the beginning of each request stays the same and more than 90\% of the input tokens are read from the cache. From Iteration~1 on, the harnesses of Run~1 and Run~3 remove the reasoning content of earlier turns and shorten older tool outputs before every model call. The part of the request after the system prompt and the task description therefore changes every time, and the share read from the cache falls to between 25\% and 33\%. The harness of Run~2 shortens the history only when it exceeds a threshold, and its share stays at about 90\%. Input that is not read from the cache costs more than ten times as much as input that is. The update prompt does not mention cost or caching.

In total, this paper reports about 11,500 runs. They use about 12.6 billion input tokens, 73\% of them read from the cache, and 193 million output tokens, and take about 2,100 hours of running time. At the official prices they cost about \$11,000, of which about \$2,000 is for evolution and \$9,000 for evaluation. The user simulators and graders are other models and are not included. They cost less than \$3 per evolution run.

\input{sections/resource_table}

\input{sections/resource_eval_table}

\section{Run Outcomes}
\label{app:outcomes}

Table~\ref{tab:outcomes} gives the numbers behind Figure~\ref{fig:outcomes} and splits the errors into three causes: a request exceeds the model's input limit, a model call fails, or a conversation on $\tau^3$-Bench banking never ends properly. A failed model call is counted only when the model's own generation is the cause, for example when the same generation exceeds the provider's time limit three times in a row. Failures of the provider's service are rerun as described in Section~\ref{sec:setup}. After evolution, no model has a run in which a request exceeds the input limit.

Table~\ref{tab:attribution} attributes the score change to the three kinds of runs. For each held-out task, every run with the seed harness is paired with every run with the other harness, and the score difference of a pair is assigned to the way its lower-scoring run ended. For example, if the Seed run ends with an error and the run with the other harness passes, the point is added under errors. If the Seed run passes and the run with the other harness times out, the point is subtracted under timeouts. The three columns add up to the score difference between the two harnesses. Table~\ref{tab:transitions} uses the same pairs and shows, for each outcome under Seed, how those runs end under Iteration~5.

\input{sections/outcome_tables}

\input{sections/transition_table}

\section{Harness Evolution Across Models}
\label{app:cross-model}

\begin{figure}[!htb]
\centering
\includegraphics[width=0.75\linewidth]{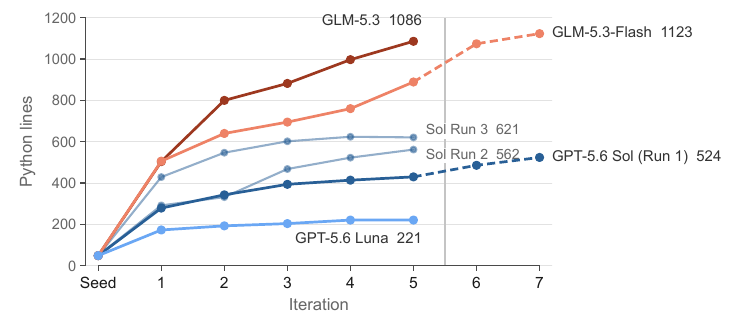}
\caption{Total Python lines of the harness after each iteration, including blank lines and comments. For GPT-5.6 Sol, the three runs of the first stage are shown. Iterations~6 and~7 (dashed) belong to the second stage, which was run with GPT-5.6 Sol, starting from Run~1, and with GLM-5.3-Flash.}
\label{fig:harness-size}
\end{figure}

\begin{figure}[!htb]
\centering
\includegraphics{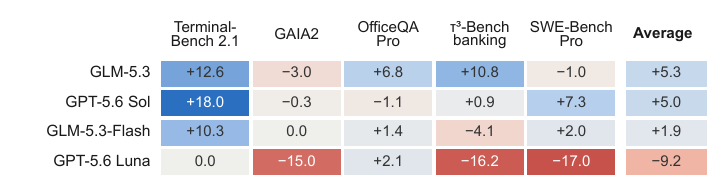}
\caption{Held-out score of Iteration~5 minus that of Seed on each in-distribution benchmark and on the average. For GPT-5.6 Sol, Iteration~5 is the mean of its three runs.}
\label{fig:gain-heatmap}
\end{figure}

Figure~\ref{fig:harness-size} shows the number of code lines of each model's harness after every iteration, with all three runs shown for GPT-5.6 Sol. Figure~\ref{fig:gain-heatmap} shows the change in score of Iteration~5 relative to Seed on each benchmark, and Table~\ref{tab:evolution-components} lists the main edits at each iteration of the three GPT-5.6 Sol runs.

The code of the two GLM models grows fastest. The three models whose scores improve all gain most on Terminal-Bench~2.1, and the drop of GPT-5.6 Luna is concentrated on GAIA2, $\tau^3$-Bench banking, and SWE-Bench~Pro.

The exploratory runs that led to the budget (Section~\ref{sec:training-view}) used GPT-5.6 Sol at max reasoning effort with no limit on added lines. In one such run on the five benchmarks, the Python code grew from 49 to 1,210 lines in five iterations, and the fourth iteration still added 232 lines. Appendix~\ref{app:batch-size} gives an example of a pattern that holds only for individual tasks.

Table~\ref{tab:budget-usage} gives the number of lines added in each iteration of the runs in this paper. At high reasoning effort, GPT-5.6 Sol adds fewer lines than in those early runs. We count the lines marked as added in the diff between two versions, over all files of the harness except the notes, so a modified line counts as one added line.

\paragraph{Model-call handling.}
The two GLM models also added more handling around the model call itself, such as retrying automatically when a single generation runs too long and is cut off by the provider, and reminding the model of the elapsed time. Such problems occur often in runs of the GLM models: the seed of GLM-5.3-Flash failed in 12 runs because a generation was cut off, and Iteration~5 has no such failure.

\paragraph{Repeated runs of the same model.}
The three independent runs of GPT-5.6 Sol evolved largely the same mechanisms. All three added the output truncation, command time limits, and history compaction of Section~\ref{sec:shared} in Iteration~1, and all three expose the tools provided by the task environment directly to the model. They differ in how individual mechanisms are implemented and in the check before finishing. When compacting the history, Run~1 deletes earlier messages, while the other two keep every message and only shorten the longer tool outputs in them. For the check before finishing, Run~1 and Run~3 both added a completion check and both tried independent review, whereas Run~2 only reminds the model once when it is about to finish without having called any tool.

\input{sections/budget_table}

\begin{table}[!htb]
\centering\small
\caption{Total Python lines of the GPT-5.6 Sol harness at each iteration of its three first-stage runs (including blank lines and comments) and the main code changes. Iterations~6 and~7 are the second stage, which continues from Run~1.}
\label{tab:evolution-components}
\renewcommand{\arraystretch}{1.16}
\begin{tabularx}{\linewidth}{@{}lrY@{}}
\toprule
\textbf{State} & \textbf{Lines} & \textbf{Main changes} \\
\midrule
Seed & 49 & System prompt, bash tool, and a basic tool-calling loop \\
\midrule
\multicolumn{3}{@{}l}{\textit{Run~1}} \\
Iteration~1 & 279 & Output truncation; command time limits; history compaction; environment tools exposed to the model directly; image viewing \\
Iteration~2 & 343 & Completion check before finishing; image size cap \\
Iteration~3 & 394 & Independent review in a fresh context \\
Iteration~4 & 414 & End-of-conversation detection by word matching; image format validation \\
Iteration~5 & 430 & Task-specific rules (set operations, deadlines); review prioritizes unchecked requirements \\
Iteration~6 & 486 & Visual inspection budget; time cap on independent review \\
Iteration~7 & 524 & Delivery deadlines for visual tasks; a tool whose description says that it ends the agent's turn ends the task \\
\midrule
\multicolumn{3}{@{}l}{\textit{Run~2}} \\
Iteration~1 & 292 & Output truncation; optional command time limit; history compaction; environment tools exposed to the model directly \\
Iteration~2 & 332 & Tool calls of one reply run in parallel; reminder when a task is about to end before any environment tool is used \\
Iteration~3 & 468 & Environment tools can ask the model for input during a call \\
Iteration~4 & 523 & Environment-tool calls of one reply run one at a time; timed-out calls are terminated \\
Iteration~5 & 562 & Command time limits enforced inside the container; very long commands shortened in the history \\
\midrule
\multicolumn{3}{@{}l}{\textit{Run~3}} \\
Iteration~1 & 429 & Output truncation; command time limits; history compaction; environment tools exposed to the model directly \\
Iteration~2 & 547 & Reminder when a task is about to end before any environment tool is used; at most one state-changing environment-tool call per reply; shorter command time limits \\
Iteration~3 & 602 & Completion check before finishing, for tasks without environment tools \\
Iteration~4 & 624 & Independent review in a fresh context; compaction keeps tool outputs until the history nears its limit \\
Iteration~5 & 621 & Independent review replaced by a check in the original conversation \\
\bottomrule
\end{tabularx}
\end{table}

\section{Mechanism Trigger Rates}
\label{app:firing}

\begin{figure}[!htb]
\centering
\includegraphics[width=\linewidth]{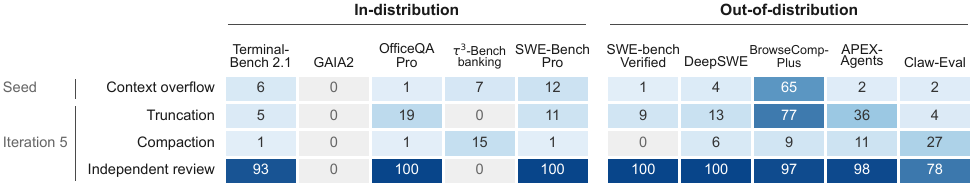}
\caption{Share of tasks (\%) on each benchmark in which the Seed of GPT-5.6 Sol aborted because of context overflow, and in which three mechanisms of its Iteration~5 from Run~1 were triggered.}
\label{fig:mechanism-firing}
\end{figure}

Figure~\ref{fig:mechanism-firing} shows how often three mechanisms of GPT-5.6 Sol are triggered on each benchmark. Seed exceeds the input limit mostly on BrowseComp-Plus. Truncation is most common in retrieval and document tasks, compaction is rare overall, and independent review runs on nearly every non-interactive task. GAIA2 and $\tau^3$-Bench banking never trigger truncation or review: their tools return short outputs, and both are handled as interactive tasks.

\section{Evolved and Production Harnesses}
\label{app:offtheshelf}

\paragraph{Shared fixes.}
Codex and Claude Code score higher than Seed mainly because they already avoid the kinds of failure that are common with the seed, and the evolved harnesses raise their scores by fixing these same kinds of failure (Figure~\ref{fig:outcomes}). Both Codex and Claude Code truncate overly long outputs, limit how long commands can run, and expose the tools provided by the task environment directly to the model. For example, on the password-recovery task of Terminal-Bench~2.1, the seed of GLM-5.3 and Claude Code ran almost the same command, and both printed about 6\,MB of output. The seed put all of it into the conversation history, the next request exceeded the input limit, and the task failed. Claude Code saved the output to a file, showed the model only a short preview, and then passed the task.

\paragraph{Remaining differences.}
For the three models whose scores improve, Iteration~5 and the production harness differ by less than 1.1 points on average, but each is ahead on some benchmarks (Table~\ref{tab:main-results}). With GPT-5.6 Sol, for example, Iteration~5 is above Codex on Terminal-Bench~2.1 and below it on GAIA2 in all three runs, because the two kinds of harness differ in how they run commands and call tools. Codex starts services inside its own terminal session, and they exit when the session ends, so it fails Terminal-Bench~2.1 tasks that require a service to keep running. The evolved harnesses take longer for every call to an environment tool and fall behind Codex on GAIA2 tasks that require actions to be completed on time.

\section{Regressions Relative to the Seed}
\label{app:regressions}

This appendix explains on which benchmarks Iteration~5 and the production harness of GPT-5.6 Sol and GLM-5.3 fall below Seed, and why (Table~\ref{tab:main-results}), and adds a few details omitted from the main text. For GPT-5.6 Sol, Iteration~5 refers to the harness from Run~1, and the scores given below are its scores.

\paragraph{GPT-5.6 Sol on $\tau^3$-Bench banking.}
Iteration~5 scores 48.65, passing 2 fewer runs than Seed, and Codex passes 4 fewer. The gap of Iteration~5 comes from history compaction. The customer's requests appear only at the start of the conversation, and in the 11 runs where Iteration~5 triggered compaction, the first compaction always deleted them. Only one of these runs passed, and in 7 the conversation never ended after compaction, until it reached the limit of 200 steps. Keeping earlier context available over a long multi-turn conversation is a known difficulty for language models~\citep{multiturnsurvey}. The failures of Codex are spread over different tasks and come from errors in business judgment and calculation. In two further runs it had correctly transferred the customer to a human agent but ended the conversation with a plain-text reply, without calling the tool that ends the conversation.

\paragraph{GPT-5.6 Sol on OfficeQA Pro.}
Iteration~5 scores 63.01 and, like Codex, is about 3 points below Seed. The questions answered wrongly are themselves ambiguous, for example whether to compute by fiscal or calendar year, or how to choose the numerator and denominator of a ratio. The same wrong answers also appear in runs of Codex and in other runs of Seed, so we find no cause for the gap in the mechanisms of the harness.

\paragraph{GLM-5.3 on GAIA2 and SWE-Bench Pro.}
On GAIA2, Iteration~5 is 3 points below Seed. It exposes the tools provided by the environment directly to the model, but the parameter descriptions of these tools have no types, and the GLM model writes integers as strings when calling them. Write operations therefore fail often, and while recovering, the model performs extra operations or omits some. Seed writes the requests itself on the command line and does not have this problem. Claude Code is 1 point below Seed on GAIA2, and Iteration~5 and Claude Code are only 1 and 2 points below Seed on SWE-Bench~Pro, all within the range of random fluctuation.

\paragraph{GPT-5.6 Sol after the second stage.}
Table~\ref{tab:retention} compares Iteration~7 and Iteration~5 of GPT-5.6 Sol on the held-out tasks of the five first-stage benchmarks. The average falls from 60.71 to 58.02 points, and the drop comes mainly from GAIA2.

The 100 held-out tasks of GAIA2 fall into five categories of 20 tasks each (Table~\ref{tab:retention-gaia2}). Almost all of the GAIA2 drop comes from the adaptability tasks, of which 8 were solved before and none is solved now. In an adaptability task, the situation changes after the agent has sent its first message to the user, and the agent has to keep working. The rule added in Iteration~7 ends the task as soon as the agent has sent a message to the user. This happened in 73 of the 100 runs, and all adaptability tasks failed as a result.

The proposer wrote this rule from the success anchors that were rerun in the last iteration, in which three of the four GAIA2 anchors failed. In one of them, the agent sent a clarification question, received no reply, and then placed an order, which cannot be undone. The description of the message tool says that a successful call ends the agent's turn. In its notes, the proposer concluded that the harness had ignored this statement, and it made the harness end the task after such a call.

The four anchors belong to the execution, ambiguity, and search categories. These three categories keep their total of 38 solved held-out tasks, whereas the adaptability and time categories, which had no anchor, fall from 9 solved tasks to 0. None of the records that the proposer read showed a task that requires more work after a message, so the rule it wrote did not consider such tasks. Replay thus protected the categories it covered but not the others. Without any replay, the proposer would see none of the requirements of the old tasks, so such conflicts would very likely be more frequent (Section~\ref{sec:claw-results}).

The amount and the coverage of the replay also matter. The GAIA2 anchors are drawn from the 28 training tasks solved in the first stage, four per iteration, which cannot cover all five categories. Replaying more tasks, sampled by task category, can reduce this problem. In addition, the rule was added in the last iteration, so no later batch tested it. Running the old tasks once more before evolution ends would catch such a case.

\input{sections/retention_table}

\input{sections/retention_gaia2_table}

\paragraph{Further details.}
In 89 of its 100 held-out runs on GAIA2, GPT-5.6 Luna saw the user's request and then deleted it when compacting the history. When GLM-5.3-Flash connected the task tools in the first stage, it attached an extra copy of every tool result, so the history reached the compaction threshold earlier. The image-viewing hint it added in the second stage also has a side effect. One Claw-Eval task asks for a scanned document to be read with a text-recognition tool. Iteration~5 called this tool and answered correctly, whereas Iteration~7 viewed the pages as images one by one until it timed out.

\section{Example Trajectories}
\label{app:traces}

This appendix gives examples of the mechanisms analyzed in Section~\ref{sec:analysis}. Each example compares two runs of the same model on the same task: one with Seed, which fails, and one with the evolved harness, which succeeds because of the mechanism. The visual inspection budget and the image-viewing tool of GLM-5.3-Flash were both added in the second stage, so these two examples compare Iteration~5 and Iteration~7. The trajectories keep only the key steps, times are minutes:seconds from the start of the run, and the steps where the mechanism takes effect and decides the outcome are marked in red.

\begin{trajbox}{Output truncation: GLM-5.3 on OfficeQA Pro uid0093}
{\small The seed of GLM-5.3 put a command output of nearly 400,000 characters into the conversation history in full. A few calls later the request exceeded the input limit, and it could not write down the answer it had already found. Iteration~5 shows the model only the truncated part and saves the full output to a file, and it answers correctly.\par}\nobreak\vspace{4pt}
\begin{trajlines}{Seed (score 0)}
00:00 & task: U.S. liabilities to the United Kingdom, CY1986, in billions -> /app/answer.txt \\
00:21-08:20 & 24 calls of grep/sed over the bulletins (CM tables, UK rows) \\
08:42 & python: list every table title that mentions "United Kingdom" in all 697 files \\
\hit{08:52} & \hit{2,608 lines (396,891 characters) are appended to the history in full} \\
\hit{09:18} & \hit{next prompt grows from 80,770 to 203,384 tokens} \\
12:23 & finds UK 1986 = 90815 in Table CM-I-3 of the June 1989 issue \\
\hit{12:25} & \hit{context\_length\_exceeded -> run aborts; answer.txt never written, score 0} \\
\end{trajlines}
\tcbline*
\begin{trajlines}{Iteration 5 (score 1)}
00:02 & harness saves the task statement to /tmp/.harness\_task.md \\
00:10-07:56 & 41 calls of grep/sed/python over the bulletins \\
\hit{22:09} & \hit{a 9-issue loop prints 61,244 characters -> about 8.9K shown, full text saved} \\
\hit{22:09} & \hit{"... use a narrower command" + path /tmp/.harness\_out\_3.txt} \\
26:02 & answer.txt: Table CM-I-3, UK end-1986 = \$90,815 million -> 90.82 \\
26:21 & finish check; answer.txt read back; peak prompt 91,651 tokens -> score 1 \\
\end{trajlines}
\end{trajbox}

\begin{trajbox}{History compaction: GPT-5.6 Sol on BrowseComp-Plus q0285}
{\small One search by the seed of GPT-5.6 Sol printed about 3.19 million characters, and the request exceeded the input limit. Iteration~5 compacted the history twice when it grew too long, reread the progress file it keeps after each compaction, continued searching, and answered correctly.\par}\nobreak\vspace{4pt}
\begin{trajlines}{Seed (score 0)}
00:00 & head -3 docids.tsv; find docs -type f | head -> corpus layout \\
00:23 & 4 parallel rg -l searches for the clues \\
00:42 & for f in <6 hits>: rg -C 4 'succeeded .*director|retir' "\$f" | head -50 \\
\hit{00:51} & \hit{3,188,315 characters come back and enter the history uncut} \\
\hit{00:54} & \hit{4th request (about 3.27M characters) -> context\_length\_exceeded -> run aborts} \\
00:54 & no answer file, score 0 \\
\end{trajlines}
\tcbline*
\begin{trajlines}{Iteration 5 (score 1)}
00:25 & rg -C 2 over 16 hits returns 1,176,087 characters \\
00:28 & model receives 45,705 characters + "[... 1130496 characters clipped by harness ...]" \\
01:10 & 47 narrower searches up to 10:11, no candidate fits yet \\
\hit{11:29} & \hit{"Older execution turns were compacted ...": history 79 -> 7 messages} \\
11:34 & cat /tmp/agent\_progress.md; cat /app/answer.txt -> neither exists yet \\
12:37 & writes the progress ledger: clues, candidates, next step \\
\hit{21:38} & \hit{second compaction (65 -> 7 messages); rereads the ledger and continues} \\
31:22 & answer.txt: Webber -> UC Citrus Experiment Station -> his student Batchelor \\
34:51 & 64 model calls; "Exact Answer: Leon Dexter Batchelor" -> correct, score 1 \\
\end{trajlines}
\end{trajbox}

\begin{trajbox}{Command time limit: GLM-5.3-Flash on Terminal-Bench 2.1 mailman}
{\small The seed of GLM-5.3-Flash waited on a command that starts a service until the task timed out. Iteration~5 terminated this command after 238 seconds, the model then started the service in the background, and the task passed.\par}\nobreak\vspace{4pt}
\begin{trajlines}{Seed (score 0)}
00:06 & cat /app/eval.py; which postfix mailman postconf -> both installed \\
03:39-15:19 & 28 calls and 42 commands reading the Mailman source and the postfix docs \\
16:21 & mailman create reading-group@local.edu -> Created mailing list \\
\hit{17:45} & \hit{mailman start 2>\&1; sleep 3; mailman status} \\
\hit{17:45-30:00} & \hit{the command never returns: 734.5 s without output, no time limit} \\
30:00 & task time limit; postfix was never started; delivery and join tests fail, score 0 \\
\end{trajlines}
\tcbline*
\begin{trajlines}{Iteration 5 (score 1)}
00:04 & ls /app; head -100 /app/eval.py; which postfix mailman postconf \\
09:34 & mailman create reading-group@local.edu -> Created mailing list \\
11:40 & postfix start -> running; SMTP probe accepted \\
12:05 & mailman start 2>\&1 | tail -5; sleep 3; mailman status \\
\hit{16:04} & \hit{harness kills the command at the 238 s limit: "run it detached instead"} \\
\hit{16:04} & \hit{harness appends a hang-watch note to this and every later tool result} \\
\hit{16:12} & \hit{setsid nohup mailman start >log 2>\&1 </dev/null \& -> GNU Mailman is running} \\
16:33-20:08 & join, confirm, post and leave tested over SMTP \\
27:11 & final message; 3 of 3 tests pass, score 1 \\
\end{trajlines}
\end{trajbox}

\begin{trajbox}{Model-call watchdog: GLM-5.3-Flash on Terminal-Bench 2.1 polyglot-c-py}
{\small The seed of GLM-5.3-Flash waited on its first model call until the task limit of 900 seconds. Iteration~5 interrupted this call after 420 seconds and issued it again, received a reply 101 seconds later, and the task passed.\par}\nobreak\vspace{4pt}
\begin{trajlines}{Seed (score 0)}
00:00 & model call 1 sent (system prompt and task); the seed waits without a deadline \\
00:00-15:00 & no reply, no tool call, nothing written \\
\hit{15:00} & \hit{call 1 is still open after 900 s -> task time limit with 0 completed calls} \\
15:00 & main.py.c does not exist, score 0 \\
\end{trajlines}
\tcbline*
\begin{trajlines}{Iteration 5 (score 1)}
00:00 & model call 1 sent; first-call watchdog set to 420 s \\
\hit{07:00} & \hit{watchdog cuts call 1 after 420 s and resends it, asking for a short reply} \\
\hit{08:41} & \hit{the retry returns in 101 s and writes main.py.c} \\
09:12 & gcc main.py.c \&\& ./cmain 0 1 2 7 10 50 -> 0 1 1 13 55 12586269025 \\
09:43 & timeout 160 apt-get install python3 -> installed \\
09:58 & python3 main.py.c 90 and ./cmain 90 -> both 2880067194370816120 \\
10:15 & final message; 1 of 1 test passes, score 1 \\
\end{trajlines}
\end{trajbox}

\begin{trajbox}{Completion check: GLM-5.3 on tau3-Bench banking task 044}
{\small Iteration~5 of GLM-5.3 twice wrote what it meant to tell the customer directly in its reply, without calling the tool that sends messages. The completion check returned both replies, and once the model sent the message with the tool, the customer applied for the new card. The seed failed on this task for a different reason: it could not state the annual fee, and the customer closed the original card.\par}\nobreak\vspace{4pt}
\begin{trajlines}{Seed (score 0)}
00:49 & customer: "I'd like to close my Gold Rewards Card ... better rewards" \\
01:02-03:24 & identity check, eligibility, card history via its own scripted tool client \\
05:33 & offers the Platinum Rewards Card -> customer: "What's the annual fee?" \\
06:47 & cannot confirm the fee -> customer: "close my Gold Rewards Card" \\
08:52 & close\_credit\_card\_account -> card closed, no application -> score 0 \\
\end{trajlines}
\tcbline*
\begin{trajlines}{Iteration 5 (score 1)}
00:13 & customer: "I'd like to close my Gold Rewards Card ... better rewards" \\
01:07-05:10 & identity check, eligibility, knowledge-base search on Rho-Bank cards \\
\hit{05:38} & \hit{recommends the Platinum card in a reply without a tool call -> finish check} \\
\hit{05:50} & \hit{send\_message\_to\_user(same text) -> customer asks about the annual fee} \\
06:21 & knowledge-base result clipped; grep in the saved copy finds the fee waiver \\
\hit{06:54} & \hit{fee answer again written as a bare reply -> finish check -> sent with the tool} \\
07:22 & customer: "I'd like to go ahead and apply" -> apply\_for\_credit\_card \\
11:14 & end\_conversation; Gold card kept, application filed -> score 1 \\
\end{trajlines}
\end{trajbox}

\begin{trajbox}{Independent review: GPT-5.6 Sol on OfficeQA Pro uid0058}
{\small Before the review, Seed and Iteration~5 reached the same wrong answer. Iteration~5 of GPT-5.6 Sol then reviewed the result independently in a new conversation, looked up the data source again, found that the figure had been revised in a later issue of the bulletin, and corrected the answer.\par}\nobreak\vspace{4pt}
\begin{trajlines}{Seed (score 0)}
00:08 & greps "Euro Positions" tables in the 2000-2001 bulletins \\
00:43 & finds the Dec-2000 row | 1962446 | 1957163 | 657186 | 619882 | in 2001 issues \\
00:54 & answer.txt: (1,962,446 - 1,957,163) + (657,186 - 619,882) = 42,587 million euros \\
\hit{00:55} & \hit{replies with the answer and no tool call -> run ends without a second look} \\
grader & expected 44174, got 42,587 -> score 0 \\
\end{trajlines}
\tcbline*
\begin{trajlines}{Iteration 5 (score 1)}
02:06 & same computation: 5,283 + 37,304 = 42,587 \\
02:31 & replies "<FINAL\_ANSWER>42,587 million euros</FINAL\_ANSWER>" with no tool call \\
\hit{02:38} & \hit{dialogue cleared: "You are the independent final reviewer ..." (task and files only)} \\
03:09 & March 2002 issue: "2000 - Dec | 1950622 | 1943752 | 657186 | 619882" \\
\hit{03:18} & \hit{the row was revised in 2002-2004 issues; python: 6,870 + 37,304 = 44,174} \\
\hit{03:52} & \hit{answer.txt <- "44,174 million euros" (latest revision)} \\
grader & expected 44174, got 44,174 -> score 1 \\
\end{trajlines}
\end{trajbox}

\begin{trajbox}{Visual inspection budget: GPT-5.6 Sol on Claw-Eval M091 (Iteration 5 vs Iteration 7)}
{\small Iteration~5 kept viewing the same frames and had written no answer when it timed out. Iteration~7 views at most 4 images per reply and never sends the same image twice, and it wrote the answer as soon as its allowance of 24 images was used up.\par}\nobreak\vspace{4pt}
\begin{trajlines}{Iteration 5 (score 0.2)}
00:00 & ffprobe: 37 s clip; 5 contact sheets; 8 image views by 00:46 \\
00:54 & history compacted; earlier images are gone \\
02:10 & after each compaction: reads the ledger, renders and views new frames \\
\hit{09:34} & \hit{13th compaction; 60 image views so far, 17 rejected as too large} \\
09:43 & requests 4 more sheets (64 views in all) \\
\hit{10:00} & \hit{600 s time limit; answer.txt never written -> score 0.2} \\
\end{trajlines}
\tcbline*
\begin{trajlines}{Iteration 7 (score 1.0)}
00:00 & one contact sheet, recompressed to fit the attachment limit \\
00:24 & per-scene sheets, then single frames, at most 4 images per turn, none resent \\
03:22 & 14 key frames in 4 turns \\
\hit{04:07} & \hit{24th image: "The visual inspection budget is exhausted. Synthesize ... and finish"} \\
\hit{04:19} & \hit{answer.txt <- four flashback scenes, each with what is on screen} \\
07:08 & final message -> score 1.0 \\
\end{trajlines}
\end{trajbox}

\begin{trajbox}{Image viewing: GLM-5.3-Flash on Claw-Eval M097 (Iteration 5 vs Iteration 7)}
{\small The task asks for the segment of a video in which two boys chase each other, delivered as a clip at double speed. Iteration~5 had no way to view images and could only estimate the position of the segment from pixel differences and motion vectors, and it had delivered nothing after 600 seconds. Iteration~7 tiled the extracted frames into contact sheets to view them, zoomed in step by step to locate the segment, and delivered in about 3 minutes.\par}\nobreak\vspace{4pt}
\begin{trajlines}{Iteration 5 (score 0.2)}
00:04 & ffprobe: 57 s video with audio \\
00:50 & frames shrunk to 64$\times$36 grey; mean pixel difference between frames \\
01:18 & differences around candidate cuts; no segment chosen \\
\hit{02:57} & \hit{motion vectors over the whole video (vidstabdetect), running for 374 s} \\
\hit{10:00} & \hit{600 s time limit; chase\_2x.mp4 and timestamp.txt never written -> score 0.2} \\
\end{trajlines}
\tcbline*
\begin{trajlines}{Iteration 7 (score 1.0)}
00:22 & one frame every 2 s with its timestamp, tiled into 2 contact sheets \\
\hit{00:38} & \hit{view\_image on both sheets} \\
00:58 & frames of 13--27 s at 1 fps; \hit{view\_image} on the new sheet \\
01:53 & frames of 15--24 s at 2 fps; \hit{view\_image} on two close-up sheets \\
\hit{03:13} & \hit{timestamp.txt <- "16 - 23"; chase\_2x.mp4 cut and sped up 2x} \\
06:52 & final message after checking the first and last frames -> score 1.0 \\
\end{trajlines}
\end{trajbox}

\section{Effect of Batch Size}
\label{app:batch-size}

Early in this project, we ran exploratory experiments on how batch size affects evolution, with one run per configuration unless stated otherwise. Batch size is the number of run records the proposer reads in one iteration. For the same total number of task runs, a larger batch means fewer iterations. The evolution runs in this appendix use GPT-5.6 Sol or GPT-5.6 Luna at max reasoning effort. When batches are too small, updates tend to fit the details of the current batch. When they are too large, there are too few updates. On Terminal-Bench~2.1, running each task one more time narrows the spread between runs.

\input{sections/batch_size_single}

\input{sections/batch_size_mixed}

\input{sections/batch_size_v2}

\paragraph{Small batches overfit.}
Each evolution run in Table~\ref{tab:batch-size} uses a single benchmark and 120 task runs in total. On GAIA2, the smaller the batch, the higher the score on the training tasks, which rises from 40.00 at batch size 120 to 55.00 at batch size 12, yet the held-out score is highest at batch size 24. The run with batch size 12 wrote the text and reference answers of training tasks into its code, and its held-out score is only 3 points above Seed. The run with batch size 24 learned the format of replies and how to write tool arguments, and its held-out score improves by 10 points.

When tasks from two benchmarks are mixed in one batch (Table~\ref{tab:batch-size-mixed}), only half of each batch consists of GAIA2 runs, and small batches fit individual tasks even more easily. The held-out score on GAIA2 is below Seed at batch size 12, and there is no such drop at batch sizes 60 and 120. For example, the run with batch size 12 turned the clarification question for the user into a fixed sentence and forbade listing the candidates. It solves only 3 of the 20 held-out tasks that need clarification, whereas the run with the same batch size that evolves on GAIA2 alone solves 14. At batch size 24 the score in Table~\ref{tab:batch-size-mixed} is also below Seed, but three later runs with this batch size score 42 to 45, close to Seed (Table~\ref{tab:batch-size-v2}). The held-out score on Terminal-Bench~2.1, in contrast, improves by 6 to 10 points at every batch size.

\paragraph{Large batches leave too few updates.}
With GPT-5.6 Luna on Terminal-Bench~2.1, every run with two or more iterations raises the held-out score to 62.07, while the run with a single iteration hardly improves. On GAIA2, the run records at batch size 120 exceeded the model's input limit, so the only iteration could not read them, and the held-out score is about the same as with Seed. GPT-5.6 Sol on Terminal-Bench~2.1 is the exception: every run improves on Seed by 6 to 15 points, and the run with a single iteration scores highest. The seed's problems on this benchmark are mainly overly long outputs and commands that never end, which a stronger model can fix in one iteration.

\paragraph{Running each task twice.}
Table~\ref{tab:batch-size-v2} compares configurations in which each task is run once or twice. With 12 Terminal-Bench~2.1 tasks and 12 GAIA2 tasks per batch, each configuration was evolved from the seed three times. The three evolution runs in which each task is run twice all reach about 84 on the Terminal-Bench~2.1 held-out tasks, above the three in which each task is run once (75 to 82), and the spread among the three shrinks from 7 points to 1. This is consistent with larger batches reducing the noise of an update. On GAIA2 there is no consistent direction, and the three evolution runs of one configuration differ by up to 13 points. With 30 GAIA2 tasks per batch, running each task once or twice gives the same score on Terminal-Bench~2.1, and both are below Seed on the GAIA2 held-out tasks.

%% file: figures/training_view.tex
\begin{figure}[!htb]
\centering
\includegraphics[width=\linewidth]{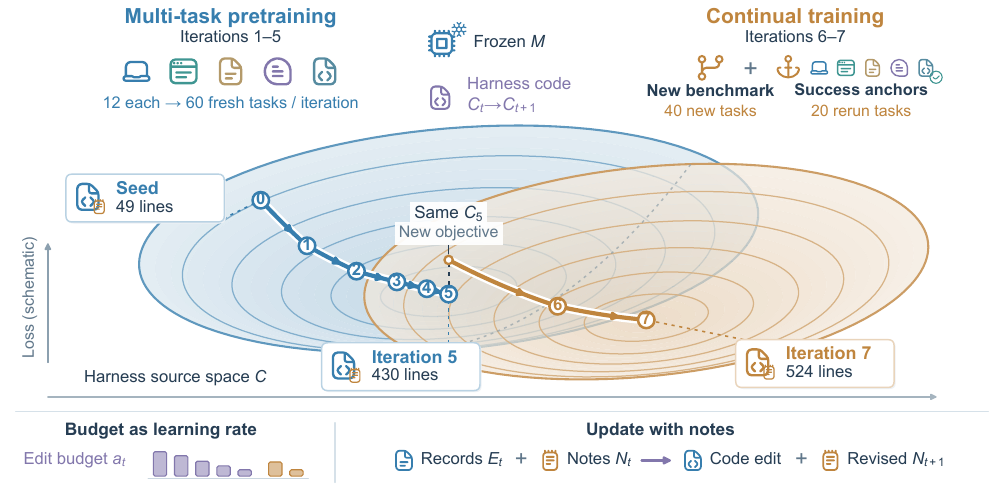}
\caption{Harness evolution viewed as two-stage training. In the first stage (multi-task pretraining, Iterations~1--5), each batch takes 12 new tasks from each of five benchmarks, 60 tasks in total. In the second stage (continual training, Iterations~6--7), each batch combines 40 tasks from a new benchmark with 20 success anchors from the first stage, rerun with the current harness. The shaded surfaces sketch the loss of the two objectives over the space of harness code, and the dotted grey curve marks where they meet. After Iteration~5 the objective changes while the code stays at $C_5$ (dashed line), and Iterations~6 and 7 continue from there. Seed, Iteration~5, and Iteration~7, the three versions evaluated on held-out tasks, have 49, 430, and 524 lines of Python. Bottom left: the budget of new lines $a_t$ shrinks during the first stage, is raised again when the second stage begins, and then shrinks again, like a learning rate schedule. Bottom right: each update turns the run records $E_t$ and the notes $N_t$ into a code edit and revised notes $N_{t+1}$. Surface shapes, step lengths, and bar heights are illustrative.}
\label{fig:training}
\end{figure}

%% file: figures/analogy_table.tex
\begin{table}[!htb]
\centering\small
\caption{Correspondence between deep learning training and harness evolution in this paper.}
\label{tab:analogy}
\renewcommand{\arraystretch}{1.0}
\begin{tabularx}{0.92\linewidth}{@{}p{4.4cm}Y@{}}
\toprule
\textbf{Deep learning} & \textbf{Harness evolution} \\
\midrule
Learnable parameters $\theta$ & All harness source code $C_t$ that runs the agent \\
Training batch & Task set $B_t$ of one iteration \\
Forward pass & The solver solves tasks with the current harness and produces trajectories \\
Loss and supervision signal & Task score $r$ and grader output $v$ \\
Backpropagation (gradient) & The proposer derives edit directions from the run records \\
Parameter update & The proposer modifies the source code, yielding $C_{t+1}$ \\
\addlinespace[0.2em]
Learning rate & Number of new lines $a_t$ allowed per iteration \\
Learning rate schedule & The budget of new lines decreases across iterations \\
Optimizer state (momentum) & Evolution notes $N_t$ kept across iterations \\
\addlinespace[0.2em]
Multi-task pretraining & Iterations~1--5: mixed tasks from five benchmarks \\
Continual training & Iterations~6--7: tasks from a new domain \\
Replay of old data & Success anchors: old tasks solved in the first stage \\
\bottomrule
\end{tabularx}
\end{table}

%% file: sections/related_table.tex
\begin{table}[!htb]
\centering\small
\caption{Comparison of recent self-evolving harness work with ours. \emph{Same model}: the proposer that modifies the harness and the solver that solves tasks use the same base model. \emph{Same harness}: the proposer itself runs on the harness being evolved. \emph{Evolution tasks}: the tasks from which one harness receives feedback during evolution. \emph{Single benchmark} means one benchmark, and \emph{same-type benchmarks} means several benchmarks or datasets of the same task type. Many methods are evaluated on several benchmarks but evolve a separate harness for each, which still counts as a single benchmark. \emph{Acceptance}: how a modified version is admitted to the next round.}
\label{tab:related}
\renewcommand{\arraystretch}{1.06}
\setlength{\tabcolsep}{3.8pt}
{\fontsize{8}{9.6}\selectfont
\begin{tabular*}{\linewidth}{@{\extracolsep{\fill}}lcccccc@{}}
\toprule
\multicolumn{1}{c}{\textbf{Method}} & \textbf{\headstack{Same\\model}} & \textbf{\headstack{Same\\harness}} & \textbf{Start} & \textbf{\headstack{Editable\\scope}} & \textbf{\headstack{Evolution\\tasks}} & \textbf{Acceptance} \\
\midrule
SICA & \yes & \yes & Hand-designed & All code & Same-type benchmarks & Score-based selection \\
DGM & \no & \yes & Minimal & All code & Single benchmark & Score-weighted sampling \\
Meta-Harness & \no & \no & Hand-designed & All code & Same-type benchmarks & Score-based selection \\
AHE & \yes & \no & Minimal & Preset components & Single benchmark & Effect-based rollback \\
Self-Harness & \yes & \yes$^{\dagger}$ & Minimal & Preset components & Single benchmark & Regression tests \\
HarnessX & \no & \no & Hand-designed & Preset components & Single benchmark & Regression tests \\
HarnessBank & \no & \no & Hand-designed & Preset components & Single benchmark & Significance test \\
HarnessCompass & \yes & \no & Minimal & Preset components & Single benchmark & Score-based selection \\
EvolveNet & \yes & \no & Minimal & All code & Single benchmark & Per-item validation gate \\
HarnessDev & \no & \no & Minimal & All code & Same-type benchmarks & Chosen by proposer \\
ModularRSI & \yes & \yes & Hand-designed & Preset components & Same-type benchmarks & Run check and review \\
RRSI & \yes & \no & Hand-designed & All code & Single benchmark & Score-based selection \\
\midrule
\textbf{Ours} & \yes & \yes & Minimal & All code & \textbf{Mixed task types} & Runnability check only \\
\bottomrule
\end{tabular*}}
\par\smallskip{\footnotesize\raggedright $^{\dagger}$The Self-Harness paper states that its proposer is invoked by the same model under the current harness, so we mark it with \yes. According to the paper and its public code repository (\url{https://github.com/qzzqzzb/Self-Harness}), the proposer receives a piece of text and outputs a structured edit plan, which a separate program parses and applies to the harness. Our proposer runs as a coding agent on the same harness as the solver and edits files directly.\par}
\end{table}

%% file: sections/ci_table.tex
\begin{table}[!htb]
\centering\small
\caption{Differences between harnesses that are compared in the main text, in points, with 95\% confidence intervals from a paired bootstrap over tasks. For GPT-5.6 Sol, Iteration~5 is the mean of its three runs on the in-distribution benchmarks and Run~1 elsewhere.}
\label{tab:ci-comparisons}
\renewcommand{\arraystretch}{1.06}
\begin{tabular}{@{}lrl@{}}
\toprule
\textbf{Comparison} & \textbf{Difference} & \textbf{95\% interval} \\
\midrule
\multicolumn{3}{@{}l}{\textit{In-distribution benchmarks: Iteration~5 minus the production harness}} \\
GLM-5.3 (Claude Code) & +0.13 & [$-$3.81, +4.08] \\
GPT-5.6 Sol (Codex) & +1.08 & [$-$2.14, +4.48] \\
GLM-5.3-Flash (Claude Code) & +0.78 & [$-$3.05, +4.56] \\
GPT-5.6 Luna (Codex) & $-$17.53 & [$-$22.48, $-$12.62] \\
\midrule
\multicolumn{3}{@{}l}{\textit{Out-of-distribution benchmarks, GPT-5.6 Sol}} \\
Iteration~5 minus Codex & $-$0.30 & [$-$3.55, +2.83] \\
\midrule
\multicolumn{3}{@{}l}{\textit{Claw-Eval, all 100 tasks}} \\
GPT-5.6 Sol: Iteration~7 minus Iteration~5 & +2.54 & [$-$2.02, +7.09] \\
GPT-5.6 Sol: Iteration~7 minus Codex & +1.60 & [$-$3.78, +6.65] \\
GLM-5.3-Flash: Iteration~7 minus Iteration~5 & +2.81 & [$-$2.51, +8.11] \\
GLM-5.3-Flash: Iteration~7 minus Claude Code & +1.87 & [$-$3.04, +6.97] \\
\midrule
\multicolumn{3}{@{}l}{\textit{Claw-Eval, 34 multimodal tasks}} \\
GPT-5.6 Sol: Iteration~7 minus Iteration~5 & +10.42 & [+0.75, +20.28] \\
GLM-5.3-Flash: Iteration~7 minus Iteration~5 & +16.98 & [+7.72, +27.16] \\
\bottomrule
\end{tabular}
\end{table}

%% file: sections/resource_table.tex
\begin{table}[!htb]
\centering\footnotesize
\caption{Tokens, running time, and API cost of the evolution runs. \emph{Runs} are the runs of the solver on training tasks, and \emph{hours} add up the running time of all runs or sessions, many of which run in parallel. Tokens are in millions, \emph{cached} is the share of input tokens read from the provider's cache, and costs are in US dollars at the official list prices. For GLM-5.3-Flash, the sessions of the first stage include one failed update and one repair.}
\label{tab:cost}
\renewcommand{\arraystretch}{1.08}
\setlength{\tabcolsep}{2.5pt}
\begin{tabular}{@{}lrrrrrrrrrrrr@{}}
\toprule
 & \multicolumn{6}{c}{\textbf{Solving training tasks}} & \multicolumn{5}{c}{\textbf{Update sessions}} & \\
\cmidrule(lr){2-7}\cmidrule(lr){8-12}
\textbf{Model} & \textbf{Runs} & \textbf{Input} & \textbf{Cached} & \textbf{Output} & \textbf{Hours} & \textbf{Cost} & \textbf{Sessions} & \textbf{Input} & \textbf{Output} & \textbf{Hours} & \textbf{Cost} & \textbf{Total cost} \\
\midrule
\multicolumn{13}{@{}l}{\textit{First stage (Iterations~1--5)}} \\
GLM-5.3 & 300 & 447.4 & 78\% & 10.8 & 79.6 & 278 & 5 & 23.2 & 0.34 & 2.2 & 12 & 290 \\
GPT-5.6 Sol, Run~1 & 300 & 190.5 & 44\% & 2.9 & 33.5 & 622 & 5 & 6.9 & 0.09 & 0.5 & 27 & 650 \\
GPT-5.6 Sol, Run~2 & 300 & 184.3 & 91\% & 2.1 & 25.9 & 189 & 5 & 12.8 & 0.09 & 0.5 & 14 & 203 \\
GPT-5.6 Sol, Run~3 & 300 & 191.0 & 46\% & 2.4 & 29.8 & 599 & 5 & 7.2 & 0.08 & 0.5 & 25 & 624 \\
GLM-5.3-Flash & 300 & 388.5 & 86\% & 6.7 & 82.1 & 22 & 7 & 28.8 & 0.43 & 4.6 & 1.5 & 23 \\
GPT-5.6 Luna & 300 & 226.5 & 36\% & 3.1 & 33.0 & 42 & 5 & 2.5 & 0.04 & 0.2 & 0.5 & 42 \\
\midrule
\multicolumn{13}{@{}l}{\textit{Second stage (Iterations~6--7)}} \\
GPT-5.6 Sol & 119 & 35.6 & 34\% & 1.2 & 13.4 & 146 & 2 & 3.0 & 0.05 & 0.4 & 15 & 161 \\
GLM-5.3-Flash & 120 & 48.4 & 62\% & 1.3 & 22.3 & 4.3 & 2 & 3.1 & 0.09 & 1.0 & 0.3 & 4.6 \\
\bottomrule
\end{tabular}
\end{table}

%% file: sections/resource_eval_table.tex
\begin{table}[!htb]
\centering\footnotesize
\caption{Tokens, running time, and API cost of the held-out evaluation runs, in the units of Table~\ref{tab:cost}. Each row adds up the runs of all the harnesses named in it, including the repeats listed in Appendix~\ref{app:benchmarks}.}
\label{tab:cost-eval}
\renewcommand{\arraystretch}{1.06}
\setlength{\tabcolsep}{5pt}
\begin{tabular}{@{}llrrrrrr@{}}
\toprule
\textbf{Model} & \textbf{Harnesses} & \textbf{Runs} & \textbf{Input} & \textbf{Cached} & \textbf{Output} & \textbf{Hours} & \textbf{Cost} \\
\midrule
\multicolumn{8}{@{}l}{\textit{Five in-distribution benchmarks}} \\
GLM-5.3 & Seed, Iteration~5, Claude Code & 1{,}521 & 2{,}981 & 91\% & 50.6 & 492 & 1{,}297 \\
GPT-5.6 Sol & Seed, Iteration~5 (three runs), Codex & 2{,}535 & 1{,}926 & 69\% & 20.2 & 240 & 3{,}887 \\
GLM-5.3-Flash & Seed, Iteration~5, Claude Code & 1{,}521 & 2{,}382 & 72\% & 40.0 & 525 & 171 \\
GPT-5.6 Luna & Seed, Iteration~5, Codex & 1{,}521 & 1{,}748 & 61\% & 23.9 & 277 & 221 \\
\midrule
\multicolumn{8}{@{}l}{\textit{Five out-of-distribution benchmarks}} \\
GPT-5.6 Sol & Seed, Iteration~5, Codex & 1{,}359 & 1{,}347 & 74\% & 16.2 & 163 & 2{,}456 \\
\midrule
\multicolumn{8}{@{}l}{\textit{Claw-Eval after the second stage}} \\
GPT-5.6 Sol & Iteration~7 & 100 & 17 & 25\% & 0.9 & 8.0 & 85 \\
GLM-5.3-Flash & Seed, Iteration~5, Iteration~7, Claude Code & 400 & 99 & 61\% & 4.6 & 48.4 & 9.8 \\
\midrule
\multicolumn{8}{@{}l}{\textit{First-stage benchmarks after the second stage}} \\
GPT-5.6 Sol & Iteration~7 & 507 & 257 & 29\% & 4.8 & 59.9 & 1{,}039 \\
\bottomrule
\end{tabular}
\end{table}

%% file: sections/outcome_tables.tex
\begin{table}[!htb]
\centering\small
\caption{How held-out runs end under each harness, as a share of runs (\%). These are the numbers behind Figure~\ref{fig:outcomes}, with errors split by cause. A run passes when it earns the full score. For GPT-5.6 Sol, Iteration~5 is the mean of its three runs on the in-distribution benchmarks and Run~1 on the out-of-distribution benchmarks.}
\label{tab:outcomes}
\renewcommand{\arraystretch}{1.08}
\begin{tabular*}{\linewidth}{@{\extracolsep{\fill}}llrrrrrrr@{}}
\toprule
 & & \multicolumn{4}{c}{\textbf{Error}} & & & \\
\cmidrule{3-6}
\textbf{Model} & \textbf{Harness} & \headstack{\textbf{Input}\\\textbf{limit}} & \headstack{\textbf{Model}\\\textbf{call}} & \headstack{\textbf{No valid}\\\textbf{end}} & \textbf{Total} & \textbf{Timeout} & \headstack{\textbf{Wrong}\\\textbf{answer}} & \textbf{Pass} \\
\midrule
\multicolumn{9}{@{}l}{\textit{In-distribution benchmarks}} \\
\multirow{3}{*}{GLM-5.3} & Seed & 2.5 & 1.8 & 1.9 & 6.1 & 12.2 & 27.9 & 53.7 \\
 & Claude Code & 0.0 & 2.4 & 0.0 & 2.4 & 7.6 & 31.1 & 58.9 \\
 & Iteration~5 & 0.0 & 0.0 & 0.3 & 0.3 & 12.0 & 28.7 & 59.0 \\
\addlinespace[0.25em]
\multirow{3}{*}{GPT-5.6 Sol} & Seed & 5.0 & 0.7 & 1.4 & 7.0 & 3.8 & 33.5 & 55.7 \\
 & Codex & 0.0 & 0.7 & 1.1 & 1.8 & 0.4 & 38.2 & 59.6 \\
 & Iteration~5 & 0.0 & 0.7 & 1.1 & 1.8 & 0.7 & 36.9 & 60.6 \\
\addlinespace[0.25em]
\multirow{3}{*}{GLM-5.3-Flash} & Seed & 2.3 & 2.1 & 2.7 & 7.1 & 15.2 & 24.5 & 53.2 \\
 & Claude Code & 0.0 & 3.0 & 0.0 & 3.0 & 9.9 & 32.8 & 54.3 \\
 & Iteration~5 & 0.0 & 0.0 & 1.4 & 1.4 & 11.8 & 31.7 & 55.1 \\
\addlinespace[0.25em]
\multirow{3}{*}{GPT-5.6 Luna} & Seed & 3.5 & 0.3 & 0.0 & 3.8 & 5.6 & 44.1 & 46.5 \\
 & Codex & 0.0 & 0.1 & 0.0 & 0.1 & 0.9 & 44.2 & 54.8 \\
 & Iteration~5 & 0.0 & 0.3 & 1.9 & 2.2 & 30.6 & 30.0 & 37.2 \\
\midrule
\multicolumn{9}{@{}l}{\textit{Out-of-distribution benchmarks}} \\
\multirow{3}{*}{GPT-5.6 Sol} & Seed & 14.8 & 0.4 & 0.0 & 15.1 & 3.8 & 33.4 & 47.7 \\
 & Codex & 0.0 & 0.0 & 0.0 & 0.0 & 4.2 & 36.7 & 59.1 \\
 & Iteration~5 & 0.0 & 0.0 & 0.0 & 0.0 & 6.6 & 33.2 & 60.2 \\
\bottomrule
\end{tabular*}
\end{table}

\begin{table}[!htb]
\centering\small
\caption{Score change relative to Seed attributed to the three kinds of runs, in points, with 95\% confidence intervals from a paired bootstrap over tasks. The three parts add up to the net change. \emph{Recovered} is the share (\%) of the Seed runs ending with an error or a timeout whose paired run under the other harness passes. Iteration~5 of GPT-5.6 Sol is defined as in Table~\ref{tab:outcomes}.}
\label{tab:attribution}
\renewcommand{\arraystretch}{1.08}
\setlength{\tabcolsep}{2.4pt}
\begin{tabular*}{\linewidth}{@{\extracolsep{\fill}}lllllrr@{}}
\toprule
\textbf{Model} & \textbf{Harness} & \textbf{From errors} & \textbf{From timeouts} & \textbf{From wrong answers} & \textbf{Net} & \textbf{Recovered} \\
\midrule
\multicolumn{7}{@{}l}{\textit{In-distribution benchmarks}} \\
\multirow{2}{*}{GLM-5.3} & Claude Code & +1.10 {\scriptsize[+0.2, +2.2]} & +1.98 {\scriptsize[$-$0.6, +4.7]} & +2.05 {\scriptsize[$-$1.1, +5.4]} & +5.13 & 26 \\
 & Iteration~5 & +0.77 {\scriptsize[0.0, +1.7]} & +1.20 {\scriptsize[$-$0.8, +3.4]} & +3.29 {\scriptsize[+0.1, +6.6]} & +5.26 & 23 \\
\addlinespace[0.25em]
\multirow{2}{*}{GPT-5.6 Sol} & Codex & +1.74 {\scriptsize[0.0, +3.7]} & +1.97 {\scriptsize[+0.7, +3.4]} & +0.16 {\scriptsize[$-$3.3, +3.7]} & +3.87 & 41 \\
 & Iteration~5 & +2.51 {\scriptsize[+1.0, +4.4]} & +2.52 {\scriptsize[+1.2, +4.0]} & $-$0.07 {\scriptsize[$-$2.9, +2.9]} & +4.95 & 48 \\
\addlinespace[0.25em]
\multirow{2}{*}{GLM-5.3-Flash} & Claude Code & +1.32 {\scriptsize[+0.1, +2.7]} & +3.34 {\scriptsize[+1.4, +5.5]} & $-$3.52 {\scriptsize[$-$6.3, $-$0.9]} & +1.15 & 27 \\
 & Iteration~5 & +1.31 {\scriptsize[+0.5, +2.3]} & +2.78 {\scriptsize[+0.3, +5.4]} & $-$2.15 {\scriptsize[$-$5.0, +0.7]} & +1.93 & 26 \\
\addlinespace[0.25em]
\multirow{2}{*}{GPT-5.6 Luna} & Codex & +2.05 {\scriptsize[+1.0, +3.3]} & +3.13 {\scriptsize[+1.2, +5.5]} & +3.12 {\scriptsize[$-$0.4, +6.8]} & +8.30 & 58 \\
 & Iteration~5 & +0.63 {\scriptsize[$-$0.3, +1.6]} & $-$5.66 {\scriptsize[$-$8.5, $-$2.8]} & $-$4.20 {\scriptsize[$-$7.3, $-$1.2]} & $-$9.23 & 30 \\
\midrule
\multicolumn{7}{@{}l}{\textit{Out-of-distribution benchmarks}} \\
\multirow{2}{*}{GPT-5.6 Sol} & Codex & +13.19 {\scriptsize[+11.0, +15.5]} & +0.58 {\scriptsize[$-$0.3, +1.6]} & $-$1.00 {\scriptsize[$-$4.4, +2.2]} & +12.77 & 73 \\
 & Iteration~5 & +12.02 {\scriptsize[+9.9, +14.2]} & $-$0.13 {\scriptsize[$-$1.1, +0.9]} & +0.57 {\scriptsize[$-$2.4, +3.6]} & +12.47 & 65 \\
\bottomrule
\end{tabular*}
\end{table}

%% file: sections/transition_table.tex
\begin{table}[!htb]
\centering\footnotesize
\caption{Outcome under Seed (rows) against outcome under Iteration~5 (columns), as a share of all runs (\%), with runs paired as in Table~\ref{tab:attribution}. Each row adds up to the share of that outcome under Seed, and each column to its share under Iteration~5.}
\label{tab:transitions}
\renewcommand{\arraystretch}{1.0}
\setlength{\tabcolsep}{7pt}
\begin{tabular}{@{}llrrrrr@{}}
\toprule
 & & & \multicolumn{4}{c}{\textbf{Under Iteration~5}} \\
\cmidrule(l){4-7}
\textbf{Model} & \textbf{Under Seed} & \textbf{Share} & \textbf{Pass} & \textbf{Error} & \textbf{Timeout} & \textbf{Wrong answer} \\
\midrule
\multicolumn{7}{@{}l}{\textit{In-distribution benchmarks}} \\
\multirow{4}{*}{GLM-5.3} & Error & 6.1 & 0.9 & 0.0 & 1.8 & 3.5 \\
 & Timeout & 12.2 & 3.2 & 0.0 & 6.8 & 2.2 \\
 & Wrong answer & 27.9 & 7.6 & 0.1 & 1.4 & 18.7 \\
 & Pass & 53.7 & 47.2 & 0.1 & 2.0 & 4.3 \\
\addlinespace[0.15em]
\multirow{4}{*}{GPT-5.6 Sol} & Error & 7.0 & 2.6 & 1.5 & 0.1 & 2.8 \\
 & Timeout & 3.8 & 2.6 & 0.0 & 0.1 & 1.1 \\
 & Wrong answer & 33.5 & 6.8 & 0.1 & 0.5 & 26.1 \\
 & Pass & 55.7 & 48.6 & 0.1 & 0.1 & 6.9 \\
\addlinespace[0.15em]
\multirow{4}{*}{GLM-5.3-Flash} & Error & 7.1 & 1.3 & 0.8 & 1.2 & 3.8 \\
 & Timeout & 15.2 & 4.4 & 0.0 & 8.5 & 2.4 \\
 & Wrong answer & 24.5 & 4.2 & 0.5 & 0.6 & 19.2 \\
 & Pass & 53.2 & 45.2 & 0.0 & 1.6 & 6.4 \\
\addlinespace[0.15em]
\multirow{4}{*}{GPT-5.6 Luna} & Error & 3.8 & 0.9 & 0.0 & 2.1 & 0.8 \\
 & Timeout & 5.6 & 1.9 & 0.0 & 2.8 & 1.0 \\
 & Wrong answer & 44.1 & 3.7 & 2.0 & 18.1 & 20.3 \\
 & Pass & 46.5 & 30.8 & 0.3 & 7.5 & 7.9 \\
\midrule
\multicolumn{7}{@{}l}{\textit{Out-of-distribution benchmarks}} \\
\multirow{4}{*}{GPT-5.6 Sol} & Error & 15.1 & 12.0 & 0.0 & 1.0 & 2.2 \\
 & Timeout & 3.8 & 0.4 & 0.0 & 2.6 & 0.8 \\
 & Wrong answer & 33.4 & 5.2 & 0.0 & 2.6 & 25.6 \\
 & Pass & 47.7 & 42.6 & 0.0 & 0.4 & 4.7 \\
\bottomrule
\end{tabular}
\end{table}

%% file: sections/budget_table.tex
\begin{table}[!htb]
\centering\small
\caption{Lines added in each iteration, compared with the budget of new lines. Iterations~6 and~7 belong to the second stage. For GPT-5.6 Sol, the run is Run~1. $^{\dagger}$The repair session of this iteration added 68 lines under its own budget of 100, and the rerun update added 96.}
\label{tab:budget-usage}
\renewcommand{\arraystretch}{1.06}
\begin{tabular}{@{}lrrrrrrr@{}}
\toprule
& \multicolumn{7}{c}{\textbf{Iteration}} \\
\cmidrule{2-8}
 & \textbf{1} & \textbf{2} & \textbf{3} & \textbf{4} & \textbf{5} & \textbf{6} & \textbf{7} \\
\midrule
Budget & 500 & 441 & 300 & 159 & 100 & 200 & 100 \\
\midrule
GLM-5.3 & 498 & 323 & 148 & 128 & 91 & -- & -- \\
GPT-5.6 Sol & 262 & 91 & 54 & 32 & 20 & 58 & 42 \\
GLM-5.3-Flash & 496 & 159 & 94 & 100 & 96$^{\dagger}$ & 193 & 76 \\
GPT-5.6 Luna & 130 & 28 & 17 & 22 & 3 & -- & -- \\
\bottomrule
\end{tabular}
\end{table}

%% file: sections/retention_table.tex
\begin{table}[!htb]
\centering\small
\caption{Held-out scores of Iteration~5 and Iteration~7 of GPT-5.6 Sol on the five first-stage benchmarks, with the difference and its 95\% confidence interval from a paired bootstrap over tasks. Iteration~5 is the harness from Run~1 with the scores of Table~\ref{tab:main-results}.}
\label{tab:retention}
\renewcommand{\arraystretch}{1.06}
\begin{tabular}{@{}lrrl@{}}
\toprule
\textbf{Benchmark} & \textbf{Iteration~5} & \textbf{Iteration~7} & \textbf{Difference} \\
\midrule
Terminal-Bench~2.1 & 83.91 & 82.76 & $-$1.15 {\scriptsize[$-$8.05, +4.60]} \\
GAIA2 & 47.00 & 38.00 & $-$9.00 {\scriptsize[$-$17.00, $-$1.00]} \\
OfficeQA Pro & 63.01 & 65.07 & +2.05 {\scriptsize[$-$2.05, +6.16]} \\
$\tau^3$-Bench banking & 48.65 & 47.30 & $-$1.35 {\scriptsize[$-$12.16, +8.11]} \\  %
SWE-Bench~Pro & 61.00 & 57.00 & $-$4.00 {\scriptsize[$-$9.00, 0.00]} \\
\midrule
Average & 60.71 & 58.02 & $-$2.69 {\scriptsize[$-$5.89, +0.49]} \\  %
\bottomrule
\end{tabular}
\end{table}

%% file: sections/retention_gaia2_table.tex
\begin{table}[!htb]
\centering\small
\caption{GAIA2 held-out tasks solved by Iteration~5 and Iteration~7 of GPT-5.6 Sol, by task category (20 tasks each), with the number of GAIA2 success anchors of each category in the batch read by the last update.}
\label{tab:retention-gaia2}
\renewcommand{\arraystretch}{1.06}
\begin{tabular}{@{}lrrr@{}}
\toprule
\textbf{Category} & \textbf{Anchors in the last batch} & \textbf{Iteration~5} & \textbf{Iteration~7} \\
\midrule
Execution & 2 & 11 & 13 \\
Ambiguity & 1 & 8 & 8 \\
Search & 1 & 19 & 17 \\
Adaptability & 0 & 8 & 0 \\
Time & 0 & 1 & 0 \\
\bottomrule
\end{tabular}
\end{table}

%% file: sections/batch_size_single.tex
\begin{table}[!htb]
\centering\small
\caption{Final harnesses of single-benchmark evolution runs that differ in batch size, i.e., the number of task runs used by one iteration. Every evolution run uses 120 task runs in total, so each of the 60 training tasks is run twice. The highest score of each group is in bold.}
\label{tab:batch-size}
\renewcommand{\arraystretch}{1.08}
\begin{tabular}{@{}lrrr@{}}
\toprule
\textbf{Batch} & \textbf{Iterations} & \textbf{Train} & \textbf{Held-out} \\
\midrule
\multicolumn{4}{@{}l}{\textit{Terminal-Bench~2.1, GPT-5.6 Sol}} \\
Seed & -- & 76.11 & 73.56 \\
12 & 10 & 86.67 & 81.61 \\
24 & 5 & \textbf{88.33} & 83.91 \\
24 (replicate) & 5 & 80.00 & 85.06 \\
60 & 2 & 86.67 & 79.31 \\
120 & 1 & 85.00 & \textbf{88.51} \\
\midrule
\multicolumn{4}{@{}l}{\textit{Terminal-Bench~2.1, GPT-5.6 Luna}} \\
Seed & -- & 68.89 & 51.72 \\
12 & 10 & \textbf{73.33} & \textbf{62.07} \\
24 & 5 & 66.67 & \textbf{62.07} \\
60 & 2 & 70.00 & \textbf{62.07} \\
120 & 1 & 68.33 & 52.87 \\
\midrule
\multicolumn{4}{@{}l}{\textit{GAIA2, GPT-5.6 Sol}} \\
Seed & -- & 31.67 & 44.00 \\
12 & 10 & \textbf{55.00} & 47.00 \\
24 & 5 & 48.33 & \textbf{54.00} \\
60 & 2 & 40.00 & 48.00 \\
120 & 1 & 40.00 & 43.00 \\
\bottomrule
\end{tabular}
\end{table}

%% file: sections/batch_size_mixed.tex
\begin{table}[!htb]
\centering\small
\caption{Final harnesses of evolution runs that mix Terminal-Bench~2.1 and GAIA2 tasks, half of each batch from each benchmark. Every training task of both benchmarks is run once, 120 task runs in all. The highest score in each column is in bold.}
\label{tab:batch-size-mixed}
\renewcommand{\arraystretch}{1.08}
\begin{tabular}{@{}lrrrrr@{}}
\toprule
& & \multicolumn{2}{c}{\textbf{Terminal-Bench~2.1}} & \multicolumn{2}{c}{\textbf{GAIA2}} \\
\cmidrule(lr){3-4}\cmidrule(l){5-6}
\textbf{Batch} & \textbf{Iterations} & \textbf{Train} & \textbf{Held-out} & \textbf{Train} & \textbf{Held-out} \\
\midrule
Seed & -- & 76.11 & 73.56 & 31.67 & 44.00 \\
12 & 10 & 88.33 & \textbf{83.91} & \textbf{45.00} & 37.00 \\
24 & 5 & 86.67 & 81.61 & 36.67 & 41.00 \\
60 & 2 & \textbf{91.67} & 79.31 & 38.33 & \textbf{50.00} \\
120 & 1 & 85.00 & 80.46 & 40.00 & 46.00 \\
\bottomrule
\end{tabular}
\end{table}

%% file: sections/batch_size_v2.tex
\begin{table}[!htb]
\centering\small
\caption{Two-benchmark evolution runs with five iterations each, in which every task of a batch is run once or twice. Configuration gives the Terminal-Bench~2.1 and GAIA2 tasks per batch and how often each task is run. Replicates repeat a configuration from the seed and list held-out scores only. The highest score in each column is in bold.}
\label{tab:batch-size-v2}
\renewcommand{\arraystretch}{1.08}
\begin{tabular}{@{}llrrrr@{}}
\toprule
& & \multicolumn{2}{c}{\textbf{Terminal-Bench~2.1}} & \multicolumn{2}{c}{\textbf{GAIA2}} \\
\cmidrule(lr){3-4}\cmidrule(l){5-6}
\textbf{Batch} & \textbf{Configuration} & \textbf{Train} & \textbf{Held-out} & \textbf{Train} & \textbf{Held-out} \\
\midrule
Seed & -- & 76.11 & 73.56 & 31.67 & 44.00 \\
24 & 12 + 12, once & \textbf{88.33} & 74.71 & \textbf{43.33} & 45.00 \\
24 & Replicate & -- & 80.46 & -- & 45.00 \\
24 & Replicate & -- & 81.61 & -- & 42.00 \\
\addlinespace[3pt]
48 & 12 + 12, twice & 85.00 & \textbf{85.06} & 41.67 & 41.00 \\
48 & Replicate & -- & 83.91 & -- & \textbf{47.00} \\
48 & Replicate & -- & 83.91 & -- & 34.00 \\
\addlinespace[3pt]
42 & 12 + 30, once & 75.00 & 78.16 & 33.33 & 40.00 \\
84 & 12 + 30, twice & 80.00 & 78.16 & 33.33 & 34.00 \\
\bottomrule
\end{tabular}
\end{table}